\documentclass[10pt,twocolumn,letterpaper]{article}

\usepackage[pagenumbers]{iccv} % To force page numbers, e.g. for an arXiv version

\usepackage{soul}
\usepackage{booktabs}
\usepackage{subcaption}
\usepackage{amsmath}
\usepackage{multirow}

\definecolor{iccvblue}{rgb}{0.21,0.49,0.74}
\usepackage[pagebackref,breaklinks,colorlinks,allcolors=iccvblue]{hyperref}

\def\paperID{*****} % *** Enter the Paper ID here
\def\confName{ICCV}
\def\confYear{2025}

\newcommand{\ours}{HQ-SMem\xspace}

\title{\ours: Video Segmentation and Tracking Using Memory Efficient Object Embedding With Selective Update and Self-Supervised Distillation Feedback}

\author{Elham Soltani Kazemi \and Imad Eddine Toubal \and Gani Rahmon \and Jaired Collins 
\and K. Palaniappan}
\begin{document}
\maketitle
\begin{abstract}

% Problem and current solutions.
Video Object Segmentation (VOS) is %pivotal in
foundational to numerous computer vision applications, including surveillance, autonomous driving, robotics and generative video editing. However, existing VOS models often struggle with precise mask delineation, deformable objects, topologically transforming objects, tracking drift and long video sequences.
% Insight, KP can move to intro to shorten abstract
%These shortcomings are attributed to the models' coarse output masks that lack topological flexibility and their redundant, memory-inefficient processing of longer video sequences.
% Method
In this paper, we introduce \ours, for \textbf{H}igh \textbf{Q}uality video segmentation and tracking using \textbf{S}mart \textbf{Mem}ory, a novel method that enhances the performance of VOS base models by addressing these limitations. Our approach incorporates three key innovations: (i) leveraging SAM with High-Quality masks (SAM-HQ) alongside appearance-based candidate-selection to refine coarse segmentation masks, resulting in improved object boundaries; (ii) implementing a dynamic smart memory mechanism that selectively stores relevant key frames while discarding redundant ones, thereby optimizing memory usage and processing efficiency for long-term videos; and (iii) dynamically updating the appearance model to effectively handle complex topological object variations and reduce drift throughout the video. These contributions mitigate several limitations of existing VOS models including, coarse segmentations that mix-in background pixels, fixed memory update schedules, brittleness to drift and occlusions, and prompt ambiguity issues associated with SAM.
% Experiments and results
%Extensive experiments conducted on multiple state-of-the-art base trackers demonstrate that our method consistently achieves performance gains of $5\%$ in tracking quality in the VOTS dataset. 
Extensive experiments conducted on multiple public datasets and state-of-the-art base trackers demonstrate that our method consistently ranks among the top two on VOTS and VOTSt 2024 datasets.
Moreover, \ours sets new benchmarks on Long Video Dataset and LVOS, showcasing its effectiveness in challenging scenarios characterized by complex multi-object dynamics over extended temporal durations.
\vspace{-1.8em}

\end{abstract}    
\section{Introduction}
\label{sec:intro}

\begin{figure}
    \centering
    \includegraphics[width=\linewidth]{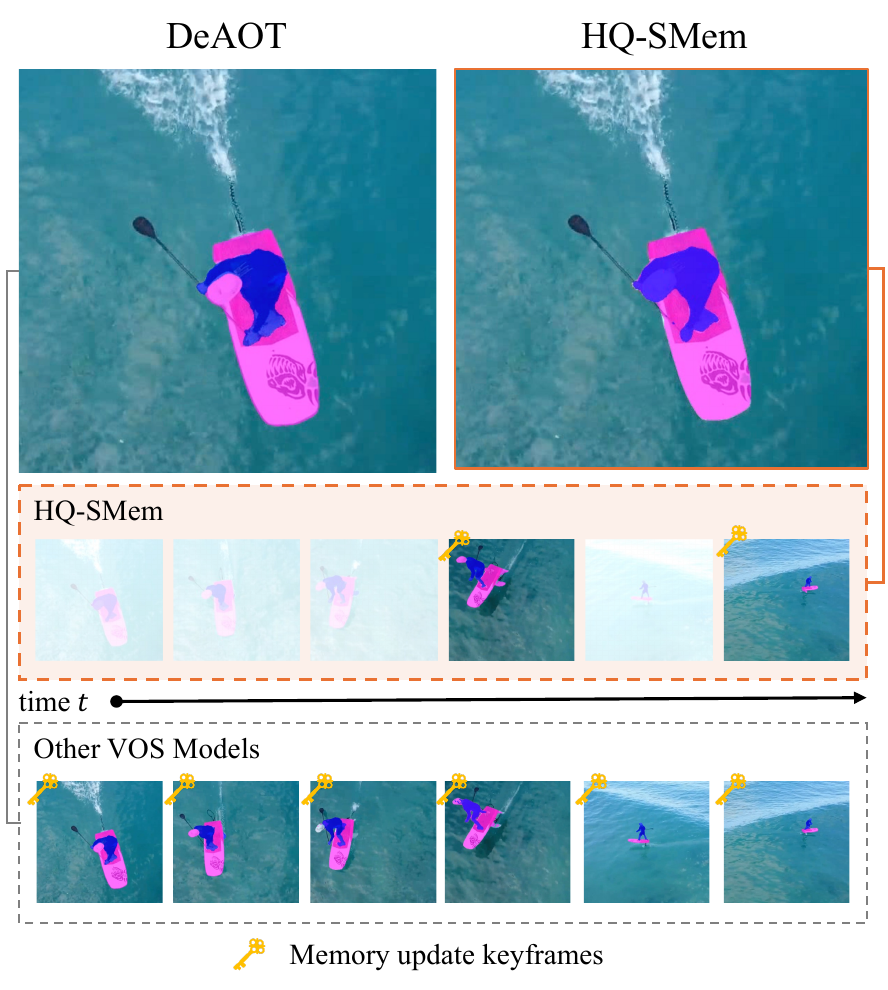}
    \caption{Our proposed \ours selectively stores only key frames that provide discriminative information for the tracker's memory, whereas DeAOT retains redundant information. In addition, the high-quality teacher-forcing mechanism enhances object boundaries, reduces object ambiguity, and provides better conditioning for the VOS encoder (see \Cref{fig:asot} for \ours architecture).
    %\hl{Add insert in upper right showing surfer with boundary of mask as yellow line. Text: HQ-SMem with selective memory update. Other VOS Models with fixed cadence memory update.}
    }
    \label{fig:figure-1}
    \vspace{-1.2em}
\end{figure}

Video Object Segmentation (VOS) is a fundamental computer vision task with applications spanning autonomous driving~\cite{siam2021video}, video editing~\cite{ravi2024sam2}, surveillance~\cite{robinson2020learning}, biomedical quantification~\cite{mavska2023cell,toubal2023ensemble} and more. The goal of VOS is to accurately segment and track multiple objects in a video, providing pixel-level masks that capture evolving object boundaries over time. Achieving precise and robust segmentation is critical for downstream applications that require fine-grained object delineation and reliable object association across video frames for perception, planning, obstacle avoidance and task execution.

Traditional approaches to VOS have ranged from basic background subtraction~\cite{mahadevan2008background,shoushtarian2005practical} to optical flow \cite{horn1981determining} to deep learning models that are driving most recent advancements in segmentation quality\cite{yang2022deaot,yang2021associating,rahmon2024deepftsg}. While methods like semantic segmentation \cite{ronneberger2015u} and instance segmentation \cite{he2017mask} have significantly improved scene understanding by performing pixel-level classification, they often lack the ability to effectively adapt to video streams. 

Building upon the advancements in transformer architectures~\cite{vaswani2017attention,dosovitskiy2021an}, end-to-end video object segmentation models such as AOT~\cite{yang2021associating} have been developed, leveraging attention mechanisms to effectively capture long-range dependencies and maintain object consistency across frames. DeAOT~\cite{yang2022deaot}, an extension of AOT, enhances video object segmentation by introducing a more efficient decoder and improving memory retrieval mechanisms for better temporal consistency. Another extension, ASOT~\cite{yang2024scalable}, further incorporates scalable long short-term transformers with layer-wise ID-based attention, enabling online architecture scalability. Numerous other methods innovated by introducing externam memory objects~\cite{cheng2022xmem} and query-based transformers~\cite{cheng2024putting}. These models tend to encounter difficulties when scaling to numerous objects, especially in longer video sequences~\cite{hong2023lvos}, where managing memory and maintaining object consistency becomes increasingly demanding. Furthermore, they often fall short in capturing fine object boundaries, particularly in scenes with intricate details or complex backgrounds.

To address these challenges, we introduce \ours, a novel method that enhances the performance of existing VOS models by leveraging three key innovations:
\begin{enumerate}
    \item The proposed \ours architecture refines VOS masks by leveraging the Segment Anything Model (SAM) and validating the outputs using DINO features to ensure high-quality video object segmentation masks reducing tracking drift.

    \item We incorporate a \textit{smart memory} management strategy that selectively stores keyframes based on significant changes in object appearance detected through DINO features, optimizing memory usage over time for better tracking in longer sequences especially with topologically-fluid object behaviors.

    \item A novel teacher-forcing mechanism is introduced to enhance object shape boundaries, reduce instance ambiguity, and provide better prompt engineering using the VOS encoder. 

\end{enumerate}

These contributions mitigate the limitations of coarse segmentations in existing VOS methods and address the ambiguity issues associated with foundation models like SAM. Our proposed \ours method has been extensively evaluated on state-of-the-art trackers across multiple datasets, demonstrating consistent performance gains in segmentation quality and setting new benchmarks on long video datasets. The results showcase the effectiveness of \ours in challenging scenarios characterized by complex object dynamics and extended temporal durations, making it a promising solution for advancing VOS in practical, real-world applications.

\begin{figure*}
    \centering
    \includegraphics[width=0.9\linewidth]{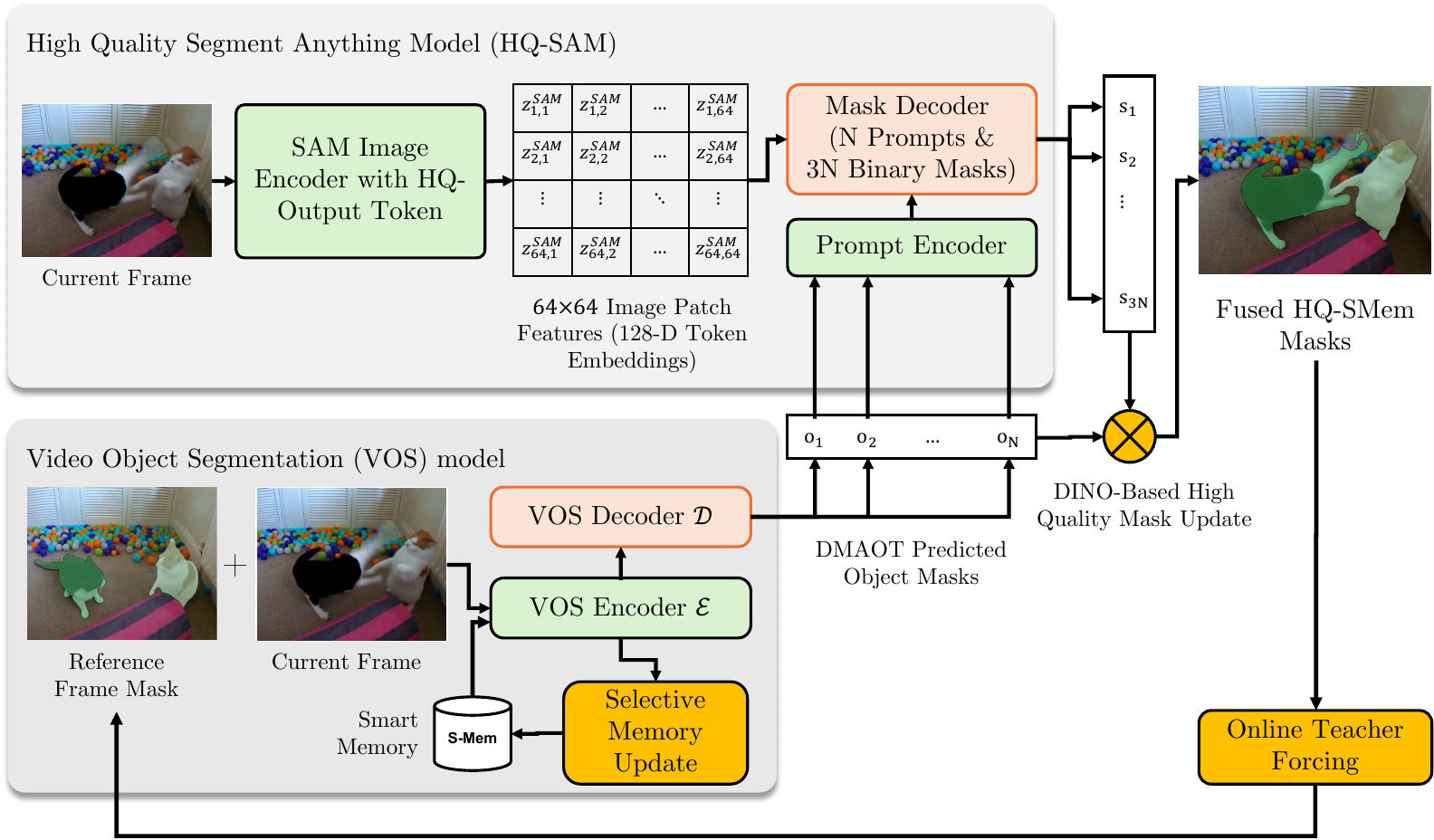}
    \caption{%ASOT system diagram. 
    System diagram for our proposed \ours architecture. The bottom-stream uses DMAOT for multi-object tracking using hierarchical propagation that is object agnostic but incorporates instant-specific information. The \emph{coarse} predicted object masks from DMAOT are used as visual prompts for HQ-SAM, in the top-stream to refine the masks with high quality detail. DINO-based fusion with teacher distillation is used
    %To avoid object-ambiguity when prompting SAM, we use an IoU-based thresholding 
    to reject SAM mask proposals that do not align with predicted objects from DMAOT.}
    \label{fig:asot}
    \vspace{-1.25em}
\end{figure*}

\section{Related Work}
\label{sec:relatedWork}

\subsection{Overview of video object segmentation (VOS)}
Video Object Segmentation (VOS) is a critical area of computer vision research dedicated to identifying and tracking objects within video frames over time. Over the last two decades, VOS has grown in importance due to its critical contribution to real-world applications. Its development has been largely propelled by deep learning advancements and benchmark datasets like the Densely Annotated Video Segmentation (DAVIS) dataset \cite{Perazzi2016, Pont-Tuset_arXiv_2017}, which set rigorous standards for evaluating segmentation performance and fostered innovation in the field \cite{Wu_2023_ICCV}. The field has transitioned from traditional methods, which struggled with occlusions and lighting inconsistencies, to more advanced neural network-based techniques that capture temporal dynamics \cite{wang2022}. Modern VOS frameworks demonstrate the integration of machine learning for enhanced segmentation accuracy and efficiency \cite{tokmakov2023breaking}. Current needs for efficient VOS algorithms that can handle scene and context-specific challenges, such as object ambiguity, appearance changes, degraded environments and topological shape changes \cite{cheng2022xmem, Zhou2024Rmem}, has led to novel solutions like XMem \cite{cheng2022xmem} and RMem \cite{Zhou2024Rmem}. %Today’s trends indicate a move toward architectures capable of managing complex, multi-object scenes in real-time, continuously pushing the boundaries of VOS capabilities.

% Video Object Segmentation (VOS) has made notable progress in recent years, motivated by the demand for efficient algorithms capable of addressing challenges inherent to video data, such as object ambiguity, changes in appearance, and sudden shape shifts \cite{cheng2022xmem, Zhou2024Rmem}. Traiditonal VOS methods have frequently faced issues with memory management and maintaining temporal consistency, prompting the development of innovative approaches like XMem \cite{cheng2022xmem} and RMem \cite{Zhou2024Rmem}.

\subsection{XMem and RMem methods in VOS}
Memory management is essential in VOS frameworks. The introduction of XMem and RMem enhances existing architectures such as AOT and DeAOT. XMem and RMem are cutting-edge approaches in Video Object Segmentation (VOS), a vital aspect of computer vision that involves tracking and segmenting objects across video frames. XMem \cite{cheng2022xmem}, introduced in 2022, features a distinct multi-store feature memory architecture, which excels at preserving object continuity, even in the presence of occlusions and complex deformations \cite{Bekuzarov2023XMemPP}. On the other hand, RMem \cite{Zhou2024Rmem}, or Restricted Memory, improves memory bank efficiency by limiting memory length during both training and inference, leading to significant boosts in VOS system performance and accuracy. Both methods address key challenges in video data, such as object ambiguity and changes in appearance, and have made a significant impact on VOS research. XMem is praised for its efficiency, supporting high frame rates while maintaining high segmentation accuracy across various video conditions, including complex camera movements and changes in object states\cite{cheng2022xmem, Bekuzarov2023XMemPP}. It seamlessly integrates with other tools, enhancing workflows in VOS applications. On the other hand, RMem tackles key limitations in traditional VOS methods by systematically managing memory banks, improving robustness and adaptability in scenarios with significant object state transitions\cite{Zhou2024Rmem}. Its plug-and-play design enables easy incorporation into existing frameworks, encouraging wider adoption in the field. Despite their advancements, both XMem and RMem still face ongoing challenges related to memory management and the adaptability of policy representations in dynamic environments. 

\subsection{DMAOT and HQ-SAM methods in VOS}
\label{sec:related:dmaot}
A recent advancement over AOT, DMAOT achieved the top rank in the VOTS 2023 challenge, showcasing its superior performance in video object tracking and segmentation. DMAOT \cite{yang2022deaot} enhances AOT \cite{yang2021associating} by employing an object-specific long-term memory strategy, enabling it to maintain consistent tracking and segmentation across frames. By utilizing memories focused on individual objects, DMAOT achieves high visual consistency for each tracked item, reducing tracking errors over time. This approach improves accuracy and stability, particularly in challenging multi-object scenarios. Despite these strengths, DMAOT typically produces coarse object masks. While sufficient for tracking purposes, these masks fall short in tasks requiring fine segmentation, potentially leading to error propagation across frames and impacting overall tracking performance. 

The runner-up method in the VOTS 2023 challenge, HQTrack \cite{hqtrack}, builds on prior winning model by integrating HQ-SAM \cite{ke2024segment} to improve mask quality. This approach enhances segmentation accuracy, overcoming limitations of previous method and boosting overall tracking performance in challenging scenarios. Building on the Segment Anything Model (SAM) \cite{kirillov2023segment}, HQ-SAM \cite{ke2024segment} offers enhanced high-quality segmentation by addressing SAM's precision limitations. It introduces a learnable High-Quality Output Token to improve resolution and detail of the mask predictions, enabling more effective handling of intricate object structures. However, HQ-SAM's visual prompting mechanism can sometimes create ambiguities, particularly when objects in the scene have similar appearances or when the prompts lack sufficient specificity.

\subsection{VOS Benchmarks}
Video Object Segmentation (VOS) has advanced significantly through various benchmarks, with DAVIS and YoutubeVOS being particularly notable. Densely Annotated VIdeo Segmentation (DAVIS) \cite{Perazzi2016, Pont-Tuset_arXiv_2017} was one of the first to improve on diversity and quality over previous benchmarks \cite{Brox2010, Fuxin2013}. It includes 50 high-quality, Full HD video sequences that capture common challenges in video object segmentation, such as occlusions, motion blur, and appearance changes. Each video is accompanied by densely annotated, pixel-accurate, per-frame ground truth segmentation. Youtube Video Object Segmentation (YoutubeVOS) \cite{YoutubeVOS2018, vos2018} expanded the scope further by collecting a much larger dataset. It consists of 4,453 high-resolution YouTube videos featuring 94 common object categories, including humans, animals, vehicles, and accessories. Each clip lasts 3-6 seconds and often contains multiple objects, which are manually segmented by professional annotators. Compared to earlier datasets, YoutubeVOS offers significantly more videos, object categories, instances, annotations, and a longer total duration of annotated videos. In this paper, we run experiments on DAVIS (2016 and 2017) and YoutubeVOS (2018 and 2019) datasets.

\subsection{DINO}
\textbf{Di}stillation with \textbf{NO} labels (DINO)~\cite{caron2021emerging}, is a self-supervised learning method for training Vision Transformer \cite{dosovitskiy2020image} models. 
It uses knowledge distillation to learn meaningful representations from images without needing labeled data. During training, DINO leverages a teacher-student framework, where the "teacher" network produces target outputs, and the "student" network learns to mimic these outputs, allowing the model to capture semantic information effectively. This approach enables DINO to perform well on various visual tasks, including object detection and segmentation, by understanding the structure and relationships within images. While DINO is primarily known for object detection, its powerful feature representations have been leveraged for tracking tasks as well \cite{tumanyan2025dino,maalouf2024follow}. In this work, we use DINO feature to associate lower quality VOS output masks with high quality SAM masks.

\section{Approach}
\label{sec:method}

Traditional VOS methods often rely on previous frame masks to update object segmentation and maintain extensive memory banks that grow linearly over time, leading to increased computational and storage demands. In this paper, we introduce a novel approach to enhance both the accuracy and efficiency of VOS through two key contributions. First, we refine the segmentation masks by using the initial VOS output as a prompt for a more advanced segmentation model, while mitigating prompt ambiguity by validating the enhanced masks through visual feature similarity; we accept the new candidate mask only if it closely aligns with the object's features. Second, we propose an adaptive memory management strategy that selectively stores key frames when significant changes in object appearance are detected using visual features, thereby optimizing memory usage without compromising segmentation performance. Our approach effectively combines high-quality mask refinement with efficient memory utilization, advancing the state of the art in video object segmentation.

\subsection{Base video object segmentation (VOS) models}

We formulate video object segmentation (VOS) models using a memory-based encoder-decoder architecture. The encoder $\mathcal{E}$ acts as a visual backbone that encodes each image frame $x_t$ into an embedding $z_t$. The decoder $\mathcal{D}$ then utilizes the current embedding $z_t$ along with a memory of previous embeddings $\mathbf{M}_{t-1} = \{z_0, z_1, \ldots, z_{t-1}\}$ to produce multi-object segmentation mask outputs $y_t = \{y_{t,1}, y_{t,2}, \ldots, y_{t,N}\}$. The function learning can be described as:
\begin{equation}
    y_t = \mathcal{D}\left(\mathcal{E}(x_t); \mathbf{M}; q\right), \text{ with } \mathbf{M} = \{\mathcal{E}(x_f)\}_{f=0}^{t-1}
    \label{eq:vos}
\end{equation}
%
%{Using} mathcal capital O may be better than lower case o which looks like zero. Could not get mathcal lower-case o to work.
where $q$ is an optional mask parameter for the decoder. In the first frame, this mask is set to the known ground-truth target object masks. In later frames, it is common for VOS models to use previous frame prediction as the optional mask parameter; i.e. $q=y_{t-1}$.

The VOS model is optimized similarly to a single-frame segmentation model by minimizing a segmentation loss function $\mathcal{L}$. Using gradient descent algorithms, we find the optimal parameters for the encoder and decoder, denoted as $\Theta_{\mathcal{E}}$ and $\Theta_{\mathcal{D}}$, respectively. The optimization objective is expressed as:
\begin{equation} \Theta_{\mathcal{E}}, \Theta_{\mathcal{D}} = \underset{\Theta_{\mathcal{E}}, \Theta_{\mathcal{D}}}{\mathbf{argmin}} \sum_{(x,g) \in \mathcal{T}} \sum_{t} \mathcal{L}\left(\mathcal{D}(\mathcal{E}(x_t); \mathbf{M}), g_t\right) \end{equation}

Here, $(x, g)$ represents a pair of input image sequences and corresponding ground-truth segmentation sequences from the training dataset $\mathcal{T}$, indexed by frame $t$. The variables $\Theta_{\mathcal{E}}$ and $\Theta_{\mathcal{D}}$ are the parameters of the encoder $\mathcal{E}$ and decoder $\mathcal{D}$ that are learned during training.
During inference, the VOS model attends to a memory bank $\mathbf{M}$ that scales with sequence length. So the space complexity of this approach is linear with video length $\mathcal{O}(T)$.

\subsection{Smart Memory (S-Mem) update}
\label{sec:smem}
In the context of VOS, the memory update problem can be framed as selecting which frame to drop from the memory bank \(\mathbf{M}\) as each new frame is processed. The majority of current SOTA approaches including \cite{yang2022deaot, cheng2022xmem, yang2021associating, hqtrack} do not use a selective memory update mechanism and consequently not only have high a memory footprint but salient appearance information in the memory bank gets diluted resulting in tracking drift. The goal is to retain the most relevant frames while discarding redundant or obsolete ones. To achieve this, we compute the relevance of each frame based on its similarity to the current frame, and select the most redundant frame to remove.

Given a memory snapshot \( \mathbf{M}_t = \{ z_1, z_2, \dots, z_{|M_t|} \} \) at time \( t \), each memory key-frame is stored as an embedding \( z_j \in \mathbb{R}^d \). We evaluate the importance of key-frames in memory using two measures that we refer to as \textit{relevance} and \textit{freshness} quantified below. 
%borrowing from the reinforcement learning literature.

\noindent \textbf{Relevance measure.}
The relevance of each past frame \( z_{j<t} \) with respect to the current frame \( z_{t} \) is measured using cosine similarity:
\begin{equation}
Rel(z_j, z_{t}) = \frac{z_j \cdot z_{t}}{\| z_j \| \| z_{t} \|} \quad \forall j < t
\end{equation}
where \( z_{t} \in \mathbb{R}^d \) is the current frame embedding.

\noindent \textbf{Freshness Measure}
The freshness of each past frame is defined as the inverse of the age of the frame \( z_j \), given by:
\begin{equation}
Fr(z_j) = \frac{1}{t - j} \quad \forall j < t
\end{equation}

As the name suggests, older frames will have a lower freshness value.

\noindent \textbf{Frame Removal Criterion}
Our selective memory strategy removes past frames that are both most relevant to the current frame (high \textit{Rel}) and older (low \textit{Fr}) compared to the other frames in the memory. Thus, we can define a removal scoring function \( O \) for each frame \( j < t \) based on both relevance and freshness defined as:
\begin{equation}
    O(z_j) = Rel(z_j, z_t) \cdot \left( 1 + \lambda \cdot Fr(z_i) \right)
\end{equation}

where \( \lambda \) is a hyper-parameter controlling the trade-off between relevance and freshness; the factor \( 1 + \lambda \) ensures that older frames are more likely to be removed when they are similar to the current frame.
To update the memory, we remove the frame that has the highest removal score:
\begin{equation}
z_{\texttt{del}} = \arg \max_{z_j} O(z_j)
\end{equation}

The updated memory \( M_{t} \) at time step \( t \) becomes:
\begin{equation}
\mathbf{M}_{t} \gets \begin{cases}
    \mathbf{M}_{t-1} \setminus \{ z_{\texttt{del}} \} \cup \{ z_{t} \},& \text{if } Rel(z_{\texttt{del}}, z_t) \geq \tau \\
    \mathbf{M}_{t-1}  \cup \{ z_{t} \}, & \text{otherwise}
\end{cases}
\end{equation}
where \( \setminus \) denotes the removal of the selected frame, and \( \cup \) denotes the addition of the current frame to the memory. $\tau$ is a threshold parameter that governs the update based on the relevance of the current feature with respect to the removed candidate.

\subsection{Online Teacher Forcing using SAM and DINO}
\label{sec:teacher_forcing}
We describe three critical synergistic enhancements in \ours that enables robustness, accuracy, scalability for VOS tracking: (i) generation of high-quality mask proposals utilizing the Segment Anything Model (SAM-HQ) \cite{ke2024segment}, (ii) validation of these mask proposals through DINO appearance features to ensure consistency with the objects being tracked by the VOS model, and (iii) implementation of online teacher forcing by reincorporating these refined masks as advanced priors into the VOS model. This process not only refines the accuracy of object segmentation but also leverages the strength of predictive modeling to enhance the continuity and quality of segmentation throughout the video.

\noindent \textbf{SAM for high-quality mask generation.} For each frame $x_{t}$, the initial VOS objects $\{y_{t, 1},y_{t, 2}, \hdots, y_{t, N}\}$ serve as a prompt to SAM to generate higher-quality masks. Using SAM's multi-mask feature, each object gets mapped into three mask proposals $\mathbf{s}_{t, i} = \{s_{t, i}^{1}, s_{t, i}^{2}, s_{t, i}^{3}\}$.

\noindent \textbf{DINO-based verification.} To address the issue of prompt ambiguity inherent in SAM, we employ DINO features to compare the VOS-predicted object with the SAM-predicted object. If the cosine similarity between their features is high, we accept the SAM output; otherwise, we retain the original VOS segmentation, and defined by:
\begin{align}
    p_{t,i} & \gets \underset{s_{i,t}^c}{\mathbf{argmax}} \ \mathbf{cos}(\phi(x_t, s_{i,t}^c), \phi(x_t, y_{t,i})) \\
    y_{t,i}^{HQ} & \gets
    \begin{cases}
        p_{t,i} & \text{if } \mathbf{cos}(\phi(x_t, p_{i,t}^c), \phi(x_t, y_{t,i})) > \tau \\
        y_{t,i} & \text{otherwise}
    \end{cases}
\end{align}
where $\phi(x, y)$ is the DINO appearance features for object $o$ in image frame $x$, and $\tau$ is the similarity threshold used for rejecting spurious masks generated due to object ambiguity. Our method is described visually in Figure \ref{fig:asot}. The final output of this step ensures a net-positive mask improvement that contributes to a higher-quality prior for the following frames. The higher quality output mask is also used as a mask prior to the next steps of the VOS base model (Eq. \eqref{eq:vos}).  This approach effectively combines the strengths of both models while mitigating potential inaccuracies due to ambiguous prompts.

% VOS models can be thought of as autoregressive models; that is, they rely on previous prediction masks in order to produce a mask for the current frame. Teacher forcing has been successfuly used in autoregressivel models in Natural Language Processing models, including LLMs. During inference, these models use previously predicted tokens to predict the next tokens. However, during training, they use ground-truth previous tokens to predict the next tokens. The idea is to have high quality priors to improve prediction. We draw parallels with our online teacher forcing design by replacing previous frames with higher quality previous frames and feeding them back to the VOS models. Factoring this in Eq. \eqref{eq:vos}, we get:
\noindent \textbf{Online Teacher Forcing.} 
VOS models inherently function as autoregressive systems, where each new prediction is dependent on prior outputs. This approach is akin to methods used in Natural Language Processing, where models are trained with teacher forcing by using ground-truth tokens to predict subsequent tokens \cite{williams1989learning}. However, during inference, the predictions from previous steps are used instead. By analogy, our method incorporates a form of online teacher forcing: we enhance the quality of prior frames by inserting high-quality masks back into the VOS pipeline. This not only ensures that subsequent frames benefit from improved initial conditions but also stabilizes and refines the overall prediction quality. Our modified equation captures this dynamic, where the VOS model utilizes high quality prior masks to generate more accurate and reliable segmentations for each new frame updating \Cref{eq:vos} to:
%This strategy effectively harmonizes the model's capabilities with the complexity of video object segmentation, ensuring each frame is characterized analytically as:
%treated with a refined analytical approach for superior results.
\begin{equation}
    y_t = \mathcal{D}\left(\mathcal{E}(x_t); \mathbf{M}_{t}; y_{t-1}^{HQ}\right)~.
\end{equation}
\section{Experiments}
\label{sec:experiments}
\subsection{Experimental setup}
\noindent \textbf{Datasets.} To comprehensively evaluate our method against state-of-the-art VOS trackers, we employ three datasets: (i) the Visual Object Tracking and Segmentation challenge (VOTS2024) dataset~\cite{kristan2023first,Kristan2024a}, which serves as the primary benchmark in the field; (ii) the Video Object Segmentation under Transformations (VOTSt) dataset~\cite{Kristan2024a}, introducing object topological transformations to assess robustness; (iii) the Long Video \cite{liang2020video} and the long-term video object segmentation (LVOS) datasets~\cite{hong2023lvos}, comprising extended sequences for evaluating memory efficiency.

\begin{figure}[h]
    \centering
    \includegraphics[width=0.8\linewidth]{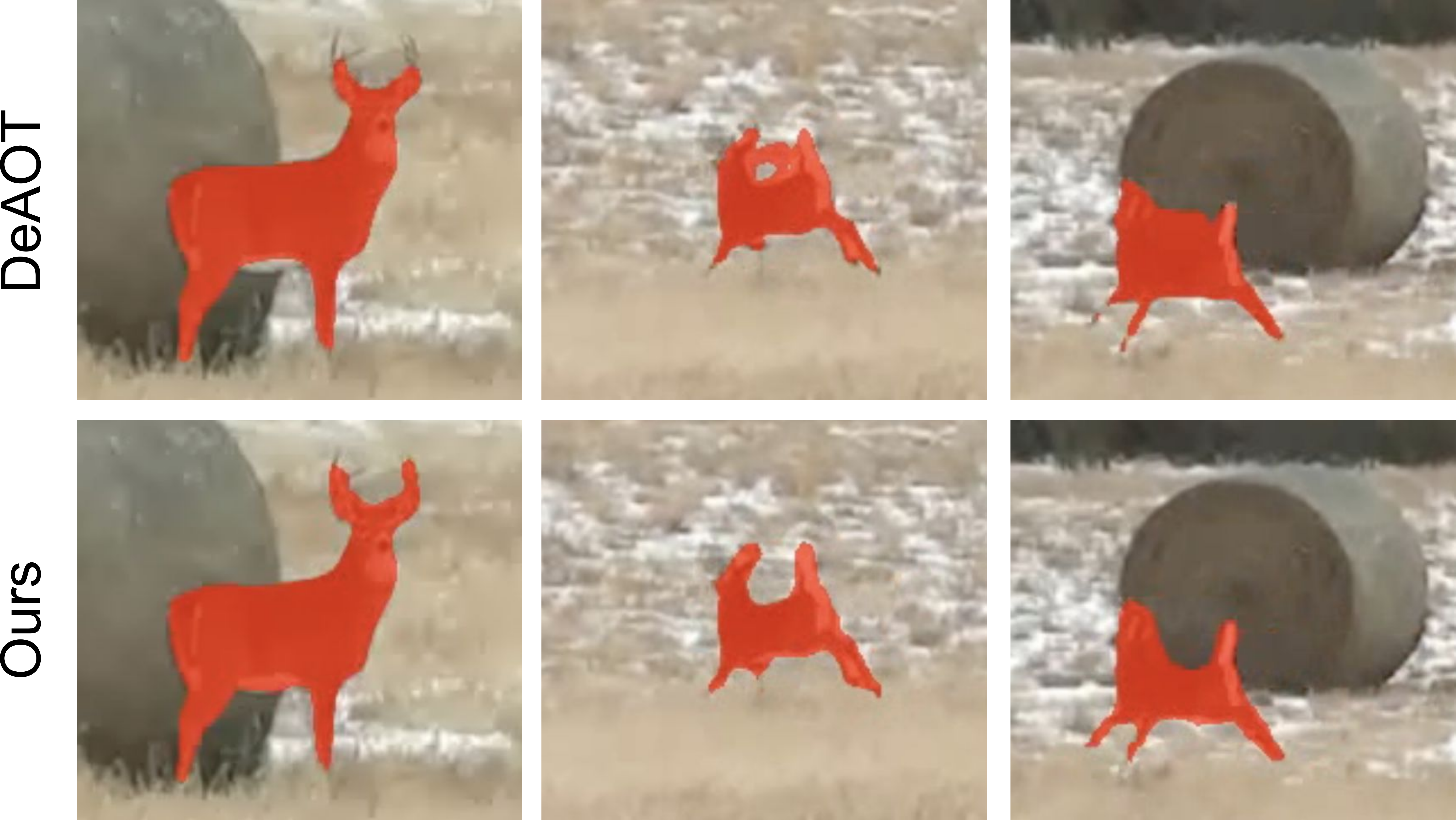}

    \vspace{2mm}
    \includegraphics[width=0.8\linewidth]{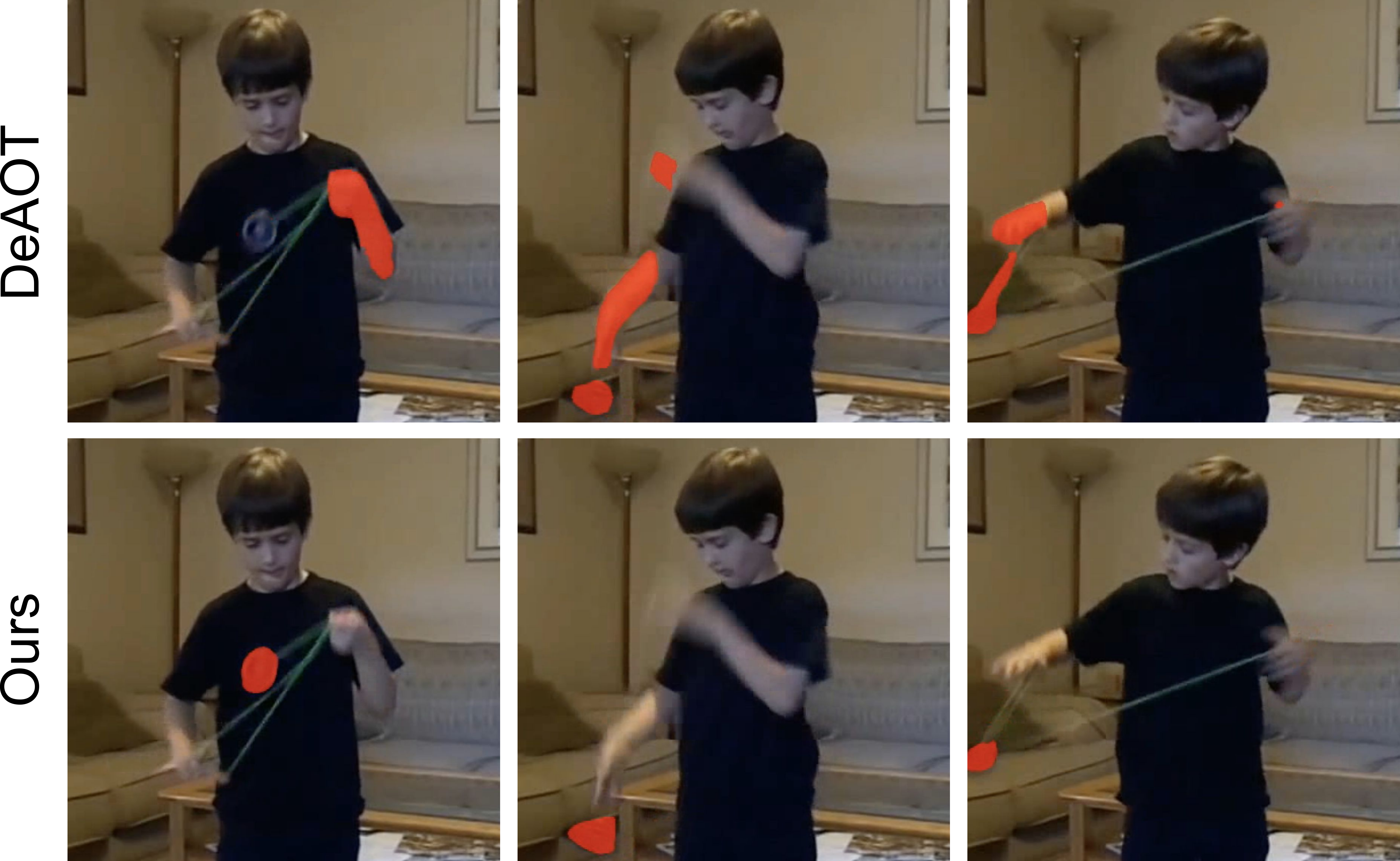}

    \vspace{2mm}
    \includegraphics[width=0.8\linewidth]{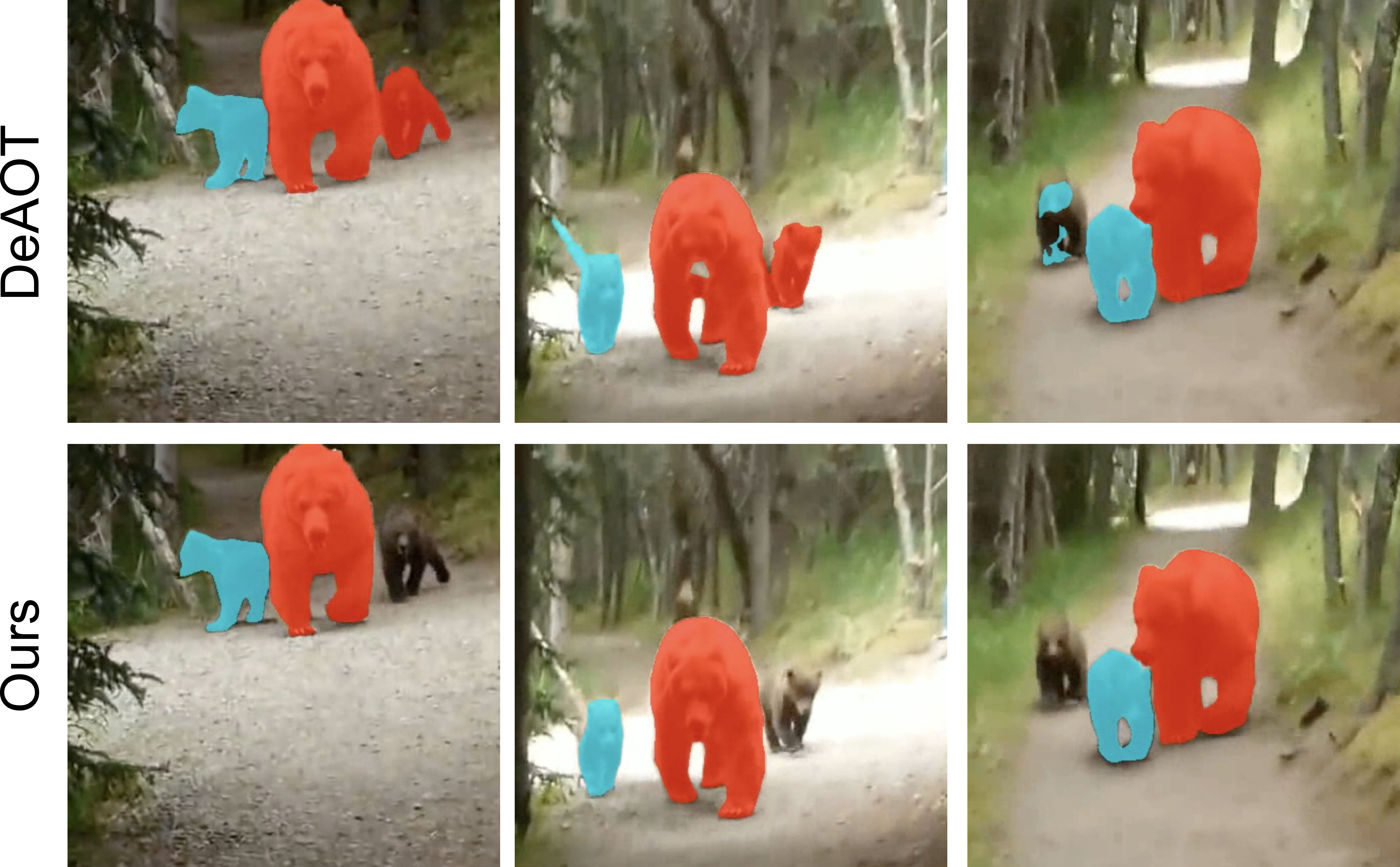}
    \caption{Comparison between DeAOT and our proposed \ours approach. \ours significantly improves the segmentation quality by enhancing mask boundaries and reducing object ambiguity, resulting in more accurate delineations and better tracking of object dynamics across frames. Our method maintains precise object boundaries and minimizes memory usage by efficiently storing key frames, achieving improved segmentation continuity and robustness, particularly in complex scenes with dynamic objects.}
    \label{fig:qualitative-vots}
\end{figure}

\noindent \textbf{Metrics.} We measure each method using the Quality metric introduced in \cite{kristan2023first, Kristan2024a}. For an entire video object segmentation  dataset of $N$ sequences  of varying frame counts $T_s$ and numbers of objects $N_s$, the quality $Q$ is given as:
\begin{equation}
    Q = \frac{1}{N}\sum_{s=1}^N \frac{1}{T_s N_s}\sum_{t=1}^{T_s}\sum_{i=1}^{N_s} \mathbf{IoU} (y_{s,t,i}, g_{s, t, i})
\end{equation}
where $y_{s,t,i}$ and $g_{s,t,i}$ is the predicted and ground-truth objects respectively for the $i^{th}$ object in frame $t$ for the sequence $s$. We show quantitative results on both VOTS 2023 and 2024 datasets \cite{kristan2023first, Kristan2024a}.

In addition to \textit{Quality (Q)}, we report several other metrics defined by the VOTS challenge~\cite{kristan2023first,Kristan2024a}: \textit{Accuracy (Acc)}, which is the sequence-normalized average overlap over successfully tracked frames; \textit{Robustness (Rob)}, representing the percentage of successfully tracked frames with the target visible, averaged over all sequences; \textit{Not-Reported Error (NRE)}, measuring the percentage of frames where the tracker incorrectly reported the target as absent; \textit{Drift-Rate Error (DRE)}, indicating the percentage of frames where the tracker drifted off the target; and \textit{Absence-Detection Quality (ADQ)}, denoting the percentage of frames where the target was correctly predicted as absent.
Finally, we describe two more metrics to measure performance on Long Video Dataset~\cite{liang2020video} and LVOS dataset~\cite{hong2023lvos}; the region similarity $\mathcal{J}$ defined in DAVIS benchmark \cite{pont20172017}.

\subsection{\ours vs dynamic object topology}

% ## INSTRUCTIONS:
% summarize table tab:votst highlighting:
% 1. \ours with DeAOT base tracker achieves state of the art performance in terms of quality, accuracy, and robustness.
% 2. \ours improves base DeAOT J and F scores by 3%
% ##

Table~\ref{tab:votst} compares the performance of \ours against state-of-the-art methods on the VOTSt dataset\cite{Kristan2024a}, which involves challenging sequences with dynamically changing object topology. Our method, \ours, using DeAOT as the base tracker, achieves state-of-the-art results across multiple metrics, including Quality (\textit{Q}), Accuracy (\textit{Acc}), and Robustness (\textit{Rob}). Notably, \ours enhances the \(\mathcal{J}\) and \(\mathcal{F}\) scores of the base DeAOT tracker by 3\%, further establishing its superiority in handling sequences where object structures undergo significant transformations.

\begin{table*}
\centering

\begin{tabular}{lllllll}
\toprule
\textbf{Method} &\textit{Q}$\uparrow$ & \textit{Acc}$\uparrow$    & \textit{Rob}$\uparrow$    & \textit{NRE}$\downarrow$  & \textit{DRE}$\downarrow$  & \textit{ADQ}$\uparrow$ \\ \midrule
S3\_Track \cite{Kristan2024a} & 0.503 & 0.499 & 0.773 & 0.083 & 0.144 & 0.287 \\ 
AOTPlus \cite{yang2021associating,Kristan2024a} & 0.485 & 0.488 & 0.725 & 0.126 & 0.150 & 0.404 \\ 
VOTST2024\_RMemAOT \cite{Kristan2024a} & 0.544 & 0.534 & 0.772 & 0.119 & 0.109 & 0.653 \\
RMem \cite{Zhou2024Rmem,Kristan2024a} & 0.515 & 0.518 & 0.777 & 0.099 & 0.125 & 0.362 \\ \hline
\textbf{\ours} (Ours) & \textbf{0.545} & \textbf{0.543} & \textbf{0.779} & 0.118 & 0.102 & 0.335 \\
\bottomrule

\end{tabular}
\caption{Performance of \ours versus state-of-the-art methods on the VOTSt dataset demonstrating our effective generalization beyond VOTS sequences to videos with objects undergoing significant topological shape changes including cutting, cooking, splitting, etc.}
\label{tab:votst}

\end{table*}

\subsection{\ours vs long videos}

Table~\ref{tab:lvd} presents a detailed comparison of \ours against state-of-the-art methods on the Long Video Dataset ~\cite{liang2020video} and LVOS dataset \cite{hong2023lvos}. \textbf{Our approach, \ours, using DeAOT as the base tracker, achieves the highest performance in terms of both the region similarity (\(\mathcal{J}\)) and contour accuracy (\(\mathcal{F}\)) metrics}, demonstrating its efficacy in handling long video sequences. Specifically, \ours improves the base DeAOT tracker's \(\mathcal{J}\) and \(\mathcal{F}\) scores by 3\%, highlighting the effectiveness of the selective memory update mechanism in preserving information over extended sequences.

\begin{table}[ht]
\centering
\begin{subtable}[t]{0.48\textwidth}
    \centering
    \caption{Long Video Dataset \cite{liang2020video}}
    \begin{tabular}{lrrr}
    \toprule
    \textbf{Method} & $\mathcal{J\&F}$ & $\mathcal{J}$ & $\mathcal{F}$ \\ \midrule
    CFBI \cite{yang2020collaborative} & 53.5 & 50.9 & 56.1 \\ 
    CFBI+ \cite{yang2021collaborative} & 50.9 & 47.9 & 53.8 \\ 
    STM \cite{oh2019video} & 80.6 & 79.9 & 81.3 \\
    MiVOS \cite{cheng2021modular} & 81.1 & 80.2 & 82.0 \\ 
    AFB-URR \cite{liang2020video} & 83.7 & 82.9 & 84.5 \\ 
    STCN \cite{cheng2021rethinking} & 87.3 & 85.4 & 89.2 \\ 
    XMem \cite{cheng2022xmem} & 89.8 & 88.0 & 91.6 \\ 
    DeAOT \cite{yang2022deaot} & 89.4 & 87.4 & 91.4 \\ 
    RMem \cite{Zhou2024Rmem}& 91.5 & 89.8 & 93.3 \\ \midrule
    DeAOT + \ours (ours) & \textbf{92.3} & \textbf{90.0} & \textbf{94.6} \\ \bottomrule
    \end{tabular}
\end{subtable}%

\vspace{5mm}

\begin{subtable}[t]{0.48\textwidth}
    \centering
    \caption{LVOS Dataset \cite{hong2023lvos}}
    \begin{tabular}{lrrr}
    \toprule
    \textbf{Method} & $\mathcal{J\&F}$ & $\mathcal{J}$ & $\mathcal{F}$ \\ \midrule
    AOT \cite{yang2021associating} & 63.6 & 57.6 & 69.5 \\ 
    DeAOT \cite{yang2022deaot} & 69.7 & 66.0 & 73.4 \\ 
    AOT + RMem \cite{Zhou2024Rmem} & 66.1 & 60.5 & 71.7 \\ \midrule
    AOT + \ours (ours) & 65.1 & 61.6 & 68.7 \\ 
    DeAOT + \ours (ours) & \textbf{73.4} & \textbf{69.7} & \textbf{77.1} \\ \bottomrule
    \end{tabular}
\end{subtable}
\caption{Performance of \ours vs. state-of-the-art methods on the Long Video Dataset ~\cite{liang2020video} and LVOS dataset \cite{hong2023lvos} demonstrating the effectiveness of our method for long-term videos.}
\label{tab:lvd}
\end{table}

\subsection{\ours and memory scaling}
\begin{figure}
    \centering
    \includegraphics[width=\linewidth]{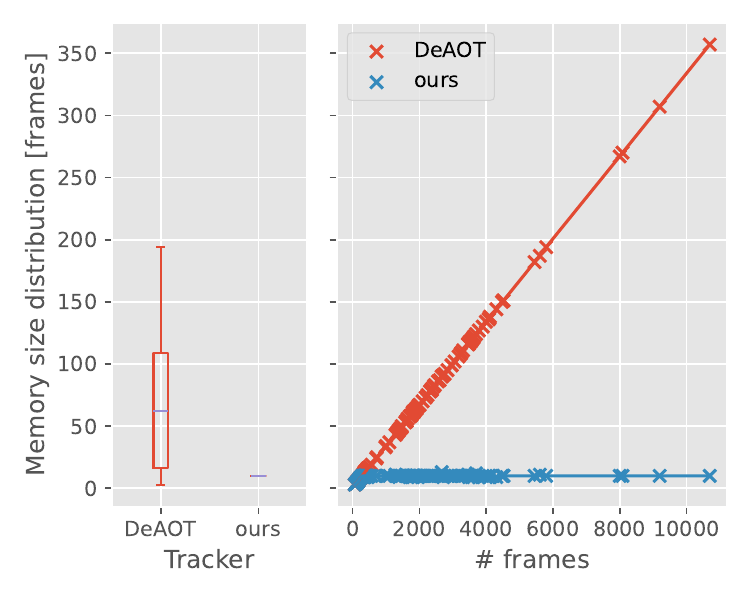}
    \caption{Memory usage per frame for DeAOT versus \ours (ours) calculated using the VOTS 2024 dataset. The box plot on the left and the line-plot on the right share the same y-axis. We are able to achieve close-to-constant scaling with a compact memory representation that also helps in improving performance of long video sequences and highly dynamic topology of objects.}
    \label{fig:memory-usage}
    \vspace{-1.0em}
\end{figure}

In order to analyze memory usage and efficiency of our method, we measure the size of the memory $|\mathbf{M}_T|$ at the end of each video sequence and plot our findings in \Cref{fig:memory-usage}.
\textbf{Our tracker demonstrates significantly better scaling compared to traditional trackers}, particularly DeAOT. As shown in \Cref{fig:memory-usage}, as the number of frames increases, our memory footprint remains consistently small (nearly flat blue line in the graph). This contrasts sharply with DeAOT, where its memory requirement increases linearly with the frame count, indicated by the positively sloped red line. This means our tracker is much more efficient in handling long video sequences, making it suitable for complex tasks that require processing extensive visual data. The box plot further emphasizes this point by showing the minimal variation in our tracker's memory size distribution compared to the wider spread for DeAOT, highlighting our consistent and compact memory representation.

\begin{table*}[ht]
\centering
\begin{tabular}{lllllll}
\toprule
  %     & \multicolumn{2}{c}{\textit{Acc \& Rob}} & & \multicolumn{3}{c}{\textit{Auxiliary measures}}   \\ \midrule
\textbf{Method}       & \textit{Q} & \textit{Acc}    & \textit{Rob}    & \textit{NRE}  & \textit{DRE}  & \textit{ADQ}   \\ \midrule
S3\_Track \cite{Kristan2024a}              & 0.722         & 0.784     & 0.889      & 0.070         & 0.041      & 0.781      \\ 
% \textbf{\ours} (Ours)         & 0.653 \footnotesize{(2)}         & 0.794 \footnotesize{(1)}      & 0.780         & 0.142         & 0.078         & 0.734          \\ 
\textbf{\ours} (Ours)         & 
0.660 & 0.793 & 0.792 & 0.139 & 0.068 &0.720\\ 
HQ-DMAOT \cite{yang2022deaot,ke2024segment}               & 0.639         & 0.754         & 0.790         & 0.138         & 0.072         & 0.750          \\ 
DMAOT \cite{yang2022deaot}                 & 0.636            & 0.751         & 0.795      & 0.139         & 0.066         & 0.731          \\ 
LY-SAM \cite{yang2022deaot,ke2024segment} & 0.631            & 0.765         & 0.776         & 0.140         & 0.084         & 0.724          \\ 
HQTrack \cite{hqtrack} & 0.615           & 0.752         & 0.766       & 0.155       & 0.079       & 0.694    \\
Cutie-SAM \cite{cheng2024putting} & 0.607            & 0.756         & 0.730         & 0.210         & 0.059      & 0.851       \\ 
oswinb\_dm\_deaot \cite{yang2022deaot,ke2024segment} & 0.597            & 0.752         & 0.845      & 0.013         & 0.141         & 0.007          \\ 
tapall\_ai \cite{cheng2024putting,radford2021learning,zagoruyko2016wide}            & 0.589            & 0.734         & 0.712         & 0.249         & 0.040      & 0.894       \\ 
goten \cite{ye2022joint,kirillov2023segment,yan2021learning}                  & 0.574            & 0.782      & 0.772         & 0.013         & 0.215         & 0.000          \\ 
AOT \cite{yang2021associating} & 0.550            & 0.698         & 0.767         & 0.096         & 0.137         & 0.470          \\ \bottomrule
\end{tabular}
\caption{%The leaderboard (top 10 trackers) from the official VOTS challenge at EECV 2024.
Top performing trackers on the official VOTS2024 challenge~\cite{Kristan2024a}.
%\hl{Add our official name from the challenge as the rank(2) result and cite Ref[14]. Then add HQ-SMem with the new performance 0.66xx}. Rankings in the top three are indicated in parentheses. 
Our \ours outperforms other SOTA methods by fusing DMAOT and high quality SAM masks using teacher forcing and smart memory (see also~Table~\ref{tab:ablation-base}). %Our integration of SAM into DMAOT shows an improvement of 2\% in Quality and 4\% in Accuracy over other implementations.
}
\vspace{-1.0em}
\label{tab:vots-leaderboard}
\end{table*}

\subsection{\ours vs dtate-of-the-art VOS methods}
We benchmark \ours against various object tracking algorithms in the VOTS challenge at EECV 2024 and summarize the results in Table~\ref{tab:vots-leaderboard}. This challenge has attracted significant attention due to its focus on real-world application scenarios. \textit{The leaderboard illustrates the competitive performance of our proposed \ours method, which would rank as the second highest.} Notably, \ours outperformed other established trackers that incorporate combinations of DMAOT and SAM technologies. 

A detailed breakdown of the scores is presented, highlighting our method's superiority in the Accuracy metric where it led with a score of 0.794, the highest among all entries. Furthermore, our method demonstrated robust improvements, enhancing Quality by 2\% and Accuracy by 4\% compared to other DMAOT and SAM integrated solutions. The superior performance of \ours underlines the benefits of our approach, particularly in handling complex tracking environments, which is critical for advancing real-time object tracking technologies.

\subsection{Qualitative analysis}
Figure \ref{fig:qualitative-vots} presents a comparative qualitative assessment of the mask quality produced by the baseline DeAOT model and our proposed HQ-SMem approach. Our method demonstrates a substantial enhancement in boundary precision and object delineation, effectively addressing the limitations observed in DeAOT. Notably, in the bear sequence, HQ-SMem produces a more accurate segmentation for the larger bear, successfully distinguishing it from the smaller, closely positioned bear, a challenge where DeAOT's masks struggled with object confusion. In the Yo-Yo sequence, HQ-SMem showcases robust tracking capabilities, maintaining clear segmentation of the yo-yo despite rapid hand movements and challenging object ambiguity. This comparison underscores HQ-SMem's effectiveness in producing fine-grained, stable masks, which is critical for high-stakes video segmentation tasks, especially in scenes involving dynamic and interacting objects.

% \begin{figure}
%     \centering
%     \includegraphics[width=\linewidth]{fig/lvos-qualitative-1.png}

%     \vspace{5mm}
%     \includegraphics[width=\linewidth]{fig/lvos-qualitative-2.png}

%     \caption{\hl{Comparison between DeAOT and our proposed approach in LVOS dataset. Can remove if we don't have VOTSt}.}
%     \label{fig:qualitative-votst}
% \end{figure}

\begin{table}[ht]
\centering
\resizebox{\linewidth}{!}{
\begin{tabular}{lcccccc}
\toprule
 & \multicolumn{2}{c}{Base} & \multicolumn{2}{c}{\ours} & \multicolumn{2}{c}{$\Delta\%\uparrow$} \\
 & $Q$ & $Acc$ & $Q$ & $Acc$ & $\Delta Q\%$ & $\Delta Acc\%$ \\
\midrule
DeAOT-T & 0.455 & 0.595 & 0.517 & 0.660 & +6.1\% & +6.4\% \\ 
DeAOT-S & 0.502 & 0.643 & 0.578 & 0.702 & +7.6\% & +5.8\% \\ 
DeAOT-B & 0.495 & 0.624 & 0.564 & 0.704 & +6.8\% & +7.9\% \\ 
DeAOT-L & 0.551 & 0.676 & 0.612 & 0.739 & +6.0\% & +6.2\% \\ 
R50\_DeAOT-L & 0.564 & 0.689 & 0.633 & 0.746 & +6.8\% & +5.7\% \\ 
Swin\_DeAOT & 0.595 & 0.739 & 0.651 & 0.789 & +5.6\% & +4.9\% \\ \midrule
DM\_DeAOT-T & 0.452 & 0.592 & 0.505 & 0.652 & +5.3\% & +6.0\% \\ 
DM\_DeAOT-S & 0.522 & 0.637 & 0.558 & 0.693 & +3.5\% & +5.5\% \\ 
DM\_DeAOT-B & 0.521 & 0.630 & 0.547 & 0.685 & +2.6\% & +5.5\% \\ 
DM\_DeAOT-L & 0.563 & 0.681 & 0.589 & 0.723 & +2.5\% & +4.2\% \\ 
DM\_R50\_DeAOT-L & 0.594 & 0.691 & 0.634 & 0.742 & +4.0\% & +5.0\% \\ 
DM\_Swin\_DeAOT & 0.636 & 0.751 & 0.660 & 0.793 & +2.4\% & +4.2\% \\
\midrule
\multicolumn{5}{r}{\textbf{Average Improvement}} & \textbf{+4.9\%} &	\textbf{+5.6\% }\\
\bottomrule
\end{tabular}
}
\caption{Ablation study of HQ-SMem with different VOS models: DeAOT and DMAOT, demonstrates a consistent improvement in both Quality (4.9\%) and Accuracy (5.6\%).}
\label{tab:ablation-base}
\end{table}

\section{Conclusions}
\ours is our innovative adaptible approach to enhance video object segmentation that addresses current limitations including coarse mask boundaries, high memory footprint, and limited perception of topologically dynamic object behaviors. \ours's combination of high-quality mask refinement, smart memory management, and online teacher for fusion, effectively bridges gaps in existing VOS tracking frameworks. Extensive experiments demonstrate \ours's competitive SOTA performance in segmentation quality, memory efficiency, and robustness across various benchmarks, including VOTS, VOTSt, Long Video Dataset and LVOS. These results highlight \ours's potential for advancing VOS in real-world intelligent perception for autonomous systems with extended temporal dynamics, tracking complex object topological changes and interactions, and dealing with degraded environments. Future work will explore using multiple smart memory selective update mechanisms, scalable memory hierarchies, optimizing for real-time performance, and extension to multimodal datasets.

\appendix

\clearpage
\setcounter{page}{1}
\setcounter{table}{0}
\setcounter{figure}{0}

\maketitlesupplementary

\section{Code}
To ensure reproducibility, we will make our code publicly available upon acceptance of the paper.

\section{Qualitative Results}
We provide a comparison video at \href{https://youtu.be/Wl-ET5oBIQg}{https://youtu.be/Wl-ET5oBIQg}, showcasing a performance evaluation of our method, \ours, against the baseline DeAOT \cite{yang2022deaot}, using a selection of videos from the VOTSt \cite{Kristan2024a}, LVOS \cite{hong2023lvos}, and VOTS \cite{Kristan2024a} datasets. We selectively picked videos representing challenging scenarios to showcase the strengths of our method. These include videos of salad and tomatoes being cut into multiple pieces from VOTSt \cite{Kristan2024a}, the sequence labeled \texttt{0tCWPOrc} for rapid motion and abrupt shape changes from LVOS \cite{hong2023lvos}, the sequence labeled \texttt{afoR2rH6} for handling objects with identical appearances and occlusions from LVOS \cite{hong2023lvos}, and \texttt{Squirrel} for its complex background and long duration from the VOTS dataset \cite{Kristan2024a}. These qualitative videos demonstrate the superior performance of our method compared to the baseline DeAOT \cite{yang2022deaot}.

\begin{figure}[h]
    \centering
    \includegraphics[width=\linewidth]{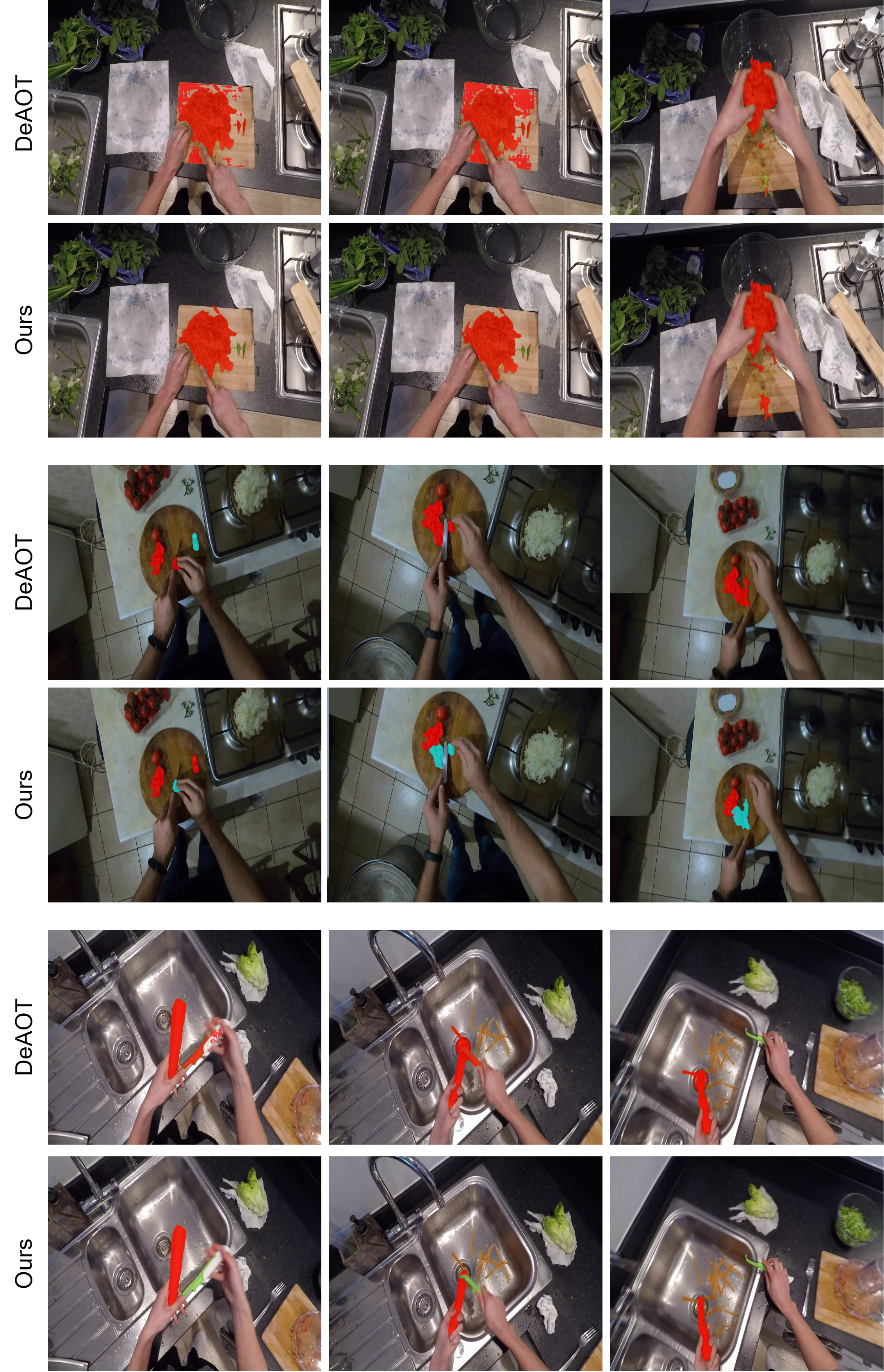}
    \caption{Comparison of DeAOT with our proposed \ours approach on three selected sequences from the VOTSt dataset. \cite{Kristan2024a}.}
    \label{fig:qualitative-votsts}
\end{figure}

\begin{figure}[h]
    \centering
    \includegraphics[width=\linewidth]{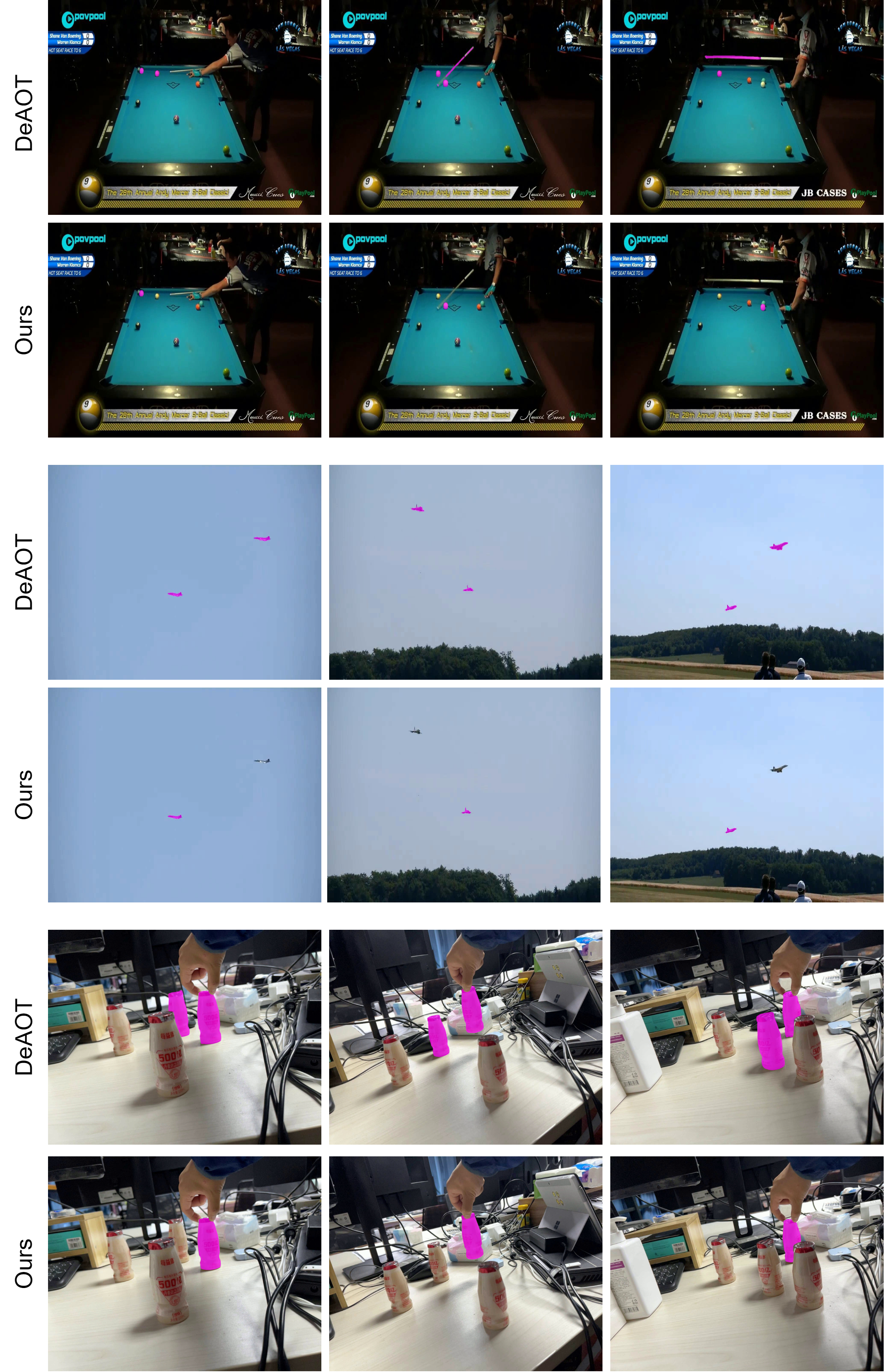}
    \caption{Comparison of DeAOT with our proposed \ours approach on selected sequences from the LVOS dataset. \cite{hong2023lvos}.}
    \label{fig:qualitative-lvos}
\end{figure}

\Cref{fig:qualitative-votsts,fig:qualitative-lvos} present qualitative comparisons between \ours and the baseline DeAOT \cite{yang2022deaot} on selected challenging sequences from the VOTSt \cite{Kristan2024a} and LVOS \cite{hong2023lvos} datasets, respectively. These figures illustrate the superior performance of \ours in handling complex scenarios such as objects being sliced into several pieces, fast motion with sudden shape transformations, identical objects experiencing occlusions, and sequences with intricate backgrounds. The visual results demonstrate that \ours consistently outperforms DeAOT, providing more accurate and reliable segmentation under these difficult conditions.

\section{Implementation Details}
In the main paper, \Cref{sec:related:dmaot,sec:method}, we present the descriptions of DMAOT/DeAOT \cite{yang2022deaot} baselines as well as our proposed approach. This section offers a comprehensive explanation of the implementation process and the configuration settings for \ours.

\subsection{Loss Functions}
Our training framework adopts the loss function formulation used in AOT \cite{yang2021associating}, DeAOT \cite{yang2022deaot}, and RMEM \cite{Zhou2024Rmem}, which combines bootstrapped cross-entropy loss with Jaccard loss. Both terms are weighted equally and averaged to compute the final loss.

\subsection{Training for VOTS, LVOS, and Long Video Dataset}
In accordance with standard practices, we first train the DeAOT model with a Swin transformer encoder \cite{yang2022deaot} on the DAVIS2017 \cite{Perazzi2016, Pont-Tuset_arXiv_2017} and YoutubeVOS2019 \cite{YoutubeVOS2018, vos2018} datasets. Subsequently, we perform inference on the VOTS \cite{Kristan2024a}, LVOS \cite{hong2023lvos}, and Long Video Dataset \cite{liang2020video}. We implement DeAOT training following the authors' original setup, using a sequence length of 5 frames, which allows for the preservation of 4 frames in the memory bank. It utilizes exponential moving averages (EMA) for parameter updates, assigning greater weight to recent values while diminishing the influence of older ones. This approach helps smooth out fluctuations by gradually incorporating new information and reducing the impact of noisy gradients. The EMA technique ensures more stable and robust training by mitigating oscillations and erratic updates, leading to more reliable convergence. 

The entire training procedure employs the AdamW optimizer and runs for 100,000 steps with a batch size of 4, distributed across 4 V100 GPUs. The initial learning rate is set to $2 \times 10^{-4}$, and it decays according to a polynomial schedule, gradually reducing to $2 \times 10^{-5}$ by the end of training. Specifically, the learning rate follows a polynomial decay function, where the rate decreases progressively over time based on:
\[
lr(t) = lr(t=0) \left(1 - \frac{t}{T}\right)^p
\]
where, $t$ represents the current step, $T$ is the total number of steps (100,000), and $p$ is the power factor controlling the rate of decay. This schedule ensures a smooth and controlled reduction in the learning rate, allowing the model to converge more effectively as training progresses. We set the learning rate of the encoder to 10\% of that of the decoder. Additionally, we employ a weight decay of 0.07, consistent with the DeAOT configuration, which serves as a regularization technique to enhance generalization and prevent the model from overfitting to the training data \cite{Zhou2024Rmem}.

\subsection{Inference for VOTS, LVOS, and Long Video Dataset}
For these challenging, and occasionally long duration, videos, we set the memory bank update frequency to $L/30$, where $L$ is the sequence length. However, by employing the updating approach outlined in \Cref{sec:smem}, we eliminate redundant information, ensuring that only distinct and useful features are retained in the memory. Additionally, we incorporate our online teacher forcing strategy, as described in \Cref{sec:teacher_forcing}, to improve tracking performance, with a particular focus on enhancing the segmentation of objects of interest. The results presented in \Cref{tab:lvd,tab:vots-leaderboard} in \Cref{sec:experiments} highlight the effectiveness of our approach.

\subsection{Training for VOTSt}
We train DeAOT \cite{yang2022deaot} on the VOST dataset \cite{tokmakov2023breaking}, using a Swin Transformer as the encoder and incorporating Temporal Positional Embeddings from RMEM \cite{Zhou2024Rmem} to enhance the model's ability to capture spatial-temporal dependencies. Subsequently, inference is performed on the VOTSt dataset \cite{Kristan2024a}. The model is initialized with pretrained weights, which were initially trained on the DAVIS2017 \cite{Perazzi2016, Pont-Tuset_arXiv_2017}  and YouTube2019 \cite{YoutubeVOS2018, vos2018} datasets. We adopt the same optimization settings as previously described, with the exception that we train for 21,000 steps instead of 100,000, and the sequence length is set to 15.

\begin{table*}
\centering

\begin{tabular}{ lll  l  l  l  l  l  l }
\toprule
VOS model & HQTF & SMem & $Q$ & $Acc$ & $Rob$ & NRE & DRE & ADQ \\
\midrule
DeAOT & & & 0.636 & 0.751 & 0.795 & 0.139 & 0.066 & 0.731 \\
DeAOT+SAM & & & 0.622 & 0.768 & 0.765 & 0.145 & 0.088 & 0.709 \\
DeAOT & & \checkmark & 0.639 & 0.758 & 0.797 & 0.140 & 0.063 & 0.759 \\
%DeAOT+SAM & 0.622 & 0.768 & 0.765 & 0.145 & 0.088 & 0.709 \\
DeAOT & \checkmark & & 0.654 & 0.788 & 0.790 & 0.139 & 0.069 & 0.721 \\
DeAOT & \checkmark & \checkmark & 0.660 & 0.793 & 0.792 & 0.139 & 0.068 & 0.720 \\
\bottomrule
\end{tabular}
\caption{Ablation study of different components in our HQ-SMem system using DeAOT as the base tracker. SAM-HQTF stands for high quality teacher forcing using SAM-HQ discussed in \Cref{sec:teacher_forcing}. S-Mem refers to Smart memory discussed in \Cref{sec:smem}. DeAOT+SAM means that the output of DeAOT is always passed to SAM, and the results from the SAM mask are used as the final output.}
\label{tab:ablation}
\end{table*}
\subsection{Inference for VOTSt}
We update the memory bank at a fixed interval of 5 frames, consistent with the update strategy used for DAVIS2017 \cite{Perazzi2016, Pont-Tuset_arXiv_2017}  and YouTubeVOS2019 \cite{YoutubeVOS2018, vos2018}. However, as with VOTS \cite{Kristan2024a}, LVOS \cite{hong2023lvos}, and Long Video Dataset \cite{liang2020video}, we apply our memory updating technique to eliminate redundant information, ensuring the memory remains compact and retains only the most relevant and representative features. Also, our online teacher forcing strategy contributes to further enhancing the final results. The results, presented in \Cref{tab:votst} in \Cref{sec:experiments}, demonstrate that our approach (\ours) outperforms RMEM \cite{Zhou2024Rmem}, S3-Track (the winner of VOST2024) \cite{Kristan2024a}, and the baseline AOT \cite{yang2021associating} by 3\%, 4.2\%, and 6\%, respectively, in terms of quality score.

\section{Ablation Study on Different Components of \ours}
We investigate how the addition of each component in \ours influences the performance of DeAOT, with a focus on two key metrics—Quality ($Q$) and Accuracy ($Acc$)—as defined in \Cref{sec:experiments}. As demonstrated in \Cref{tab:ablation}, the baseline DeAOT achieves a $Q$ score of 0.636 and an $Acc$ of 0.751 on the VOTS dataset \cite{Kristan2024a}, providing a robust baseline for object tracking performance. Introducing our Smart Memory (SMem) technique, as described in \Cref{sec:smem}, \emph{without} High-Quality Teacher Forcing (SAM-HQTF), as introduced in \Cref{sec:teacher_forcing}, yields a subtle but promising improvement, with $Q$ increasing to 0.639 and $Acc$ rising to 0.758. This indicates that SMem improves the model's capacity to retain relevant information from previous frames, thereby enhancing tracking stability and precision. The improvement though small is often on critical frames and reduces mask drifting.

Naively incorporating Segment Anything Model (SAM) \cite{kirillov2023segment} into DeAOT, by directly using the mask output from SAM without any validation, leads to a decrease in $Q$ (0.622) but an improvement in $Acc$ (0.768). This indicates that while SAM enhances the model’s focus on the target object, it may also introduce some challenges that compromise the overall tracking quality. On the other hand, when combined with SAM-HQTF both $Q$ (0.654) and $Acc$ (0.788) show substantial improvements. This demonstrates that the integration of SAM-HQTF provides enhanced supervision, allowing the model to better learn object representations and further refine its tracking performance.

Finally, the complete configuration—combining SAM-HQTF and SMem with DeAOT, as in our \ours approach—yields the best performance, achieving a $Q$ of 0.660 and an $Acc$ of 0.793. This configuration exemplifies the complementary synergy between SMem and SAM-HQTF, effectively enhancing both the robustness and precision of the tracker. Overall, these results highlight the significant contribution of each component to enhancing the model's tracking accuracy and overall performance, with the most substantial improvements achieved through the integration of SAM-HQTF and SMem.

\begin{table}
\centering
\begin{tabular}{ c r l l l }
\toprule
\multicolumn{2}{l}{\textbf{Visual Prompt}} & \multirow{2}{*}{$Q$}  & \multirow{2}{*}{$Acc$} & \multirow{2}{*}{$Rob$} \\
Box & \# Points &  &  &  \\
\midrule
\checkmark & & 0.660 & 0.793 & 0.792 \\ 
\checkmark & \texttt{5+} & 0.659 & 0.783 & 0.796 \\ 
\checkmark & \texttt{5-} & 0.656 & 0.779 & 0.795 \\ 
\checkmark & \texttt{10+} & 0.656 & 0.779 & 0.796 \\ 
\checkmark & \texttt{10-} & 0.652 & 0.773 & 0.796 \\ 
\checkmark & \texttt{15+} & 0.652 & 0.774 & 0.796 \\ 
\checkmark & \texttt{15-} & 0.649 & 0.769 & 0.796 \\ 
\checkmark & \texttt{20+} & 0.649 & 0.769 & 0.796 \\ 
\checkmark & \texttt{20-} & 0.647 & 0.766 & 0.796 \\ 
           & \texttt{5+} & 0.643 & 0.764 & 0.794 \\ 
           & \texttt{5-} & 0.617 & 0.724 & 0.796 \\ 
           & \texttt{10+} & 0.642 & 0.764 & 0.794 \\
           & \texttt{10-} & 0.616 & 0.723 & 0.797 \\ 
\bottomrule
\end{tabular}
\caption{Ablation study on visual prompting of Segment Anything Model on the test set of the VOTS challenge. Each row represents an experiment where SAM is fed a bounding box and/or a set of positive (\texttt{+}) or negative (\texttt{-}) points. Despite the options that SAM offers in terms of prompting, we found that using simply the bounding box as a visual prompt to SAM provides the best results.}
\label{tab:visual_prompt_ablation}
\end{table}

\section{Visual Prompting Ablation}
To assess the impact of different visual prompts on the performance of SAM\cite{kirillov2023segment}, we conducted an ablation study summarized in Table~\ref{tab:visual_prompt_ablation}. We experimented with various combinations of bounding boxes and sets of positive (\texttt{+}) or negative (\texttt{-}) points as inputs to SAM. The results indicate that using only the bounding box as a visual prompt yields the highest performance across all evaluated metrics, achieving a $Q$ score of 0.660, an $Acc$ of 0.793, and robustness ($Rob$) of 0.792. Incorporating additional points—whether positive or negative and in varying quantities—did not enhance performance; in fact, it often led to marginal decreases in the metrics. These findings suggest that the bounding box alone is the most effective visual prompt for SAM in our application.

\section{Discussion on Limitations and Future Work}
Using high quality prompt-based mask refinement, smart memory updating with dynamic feedback of appearance and topological changes help to establish a new SOTA on several challenging VOS benchmark datasets as described in the main paper. We point out several limitations of our current version of \ours that can be investigated to develop better approaches. One is reduce mask drift when there is complex topological change like in the videos of salad lettuce cutting and tomato cutting of VOTSt, or rapid scale changes like in the LVOS surfer video (see \Cref{fig:figure-1} in main paper) and drone video (see \Cref{fig:qualitative-lvos} middle rows). Object instance confusion can happen when there are multiple high occlusion events with complex inter-object interactions as in the LVOS container shuffling video (see  \Cref{fig:qualitative-lvos} bottom rows), billiard table striking pool ball video (see \Cref{fig:qualitative-lvos} top rows) and the VOS squirrel carrying bagel video where only the squirrel is to be segmented (see Supplementary video). \ours is currently limited in the number of distinct objects that can be tracked (around ten) due to the memory constraints of DeAOT. An advantage of our \ours is that the approach can be transferred to other VOS algorithms without the same memory constraints to scale to tracking dozens of objects. Many similar objects of small scale like farm animals (ie. sheep and cows) from an aerial view are easily confused or multiple horses and jockeys in a horse race when only a few objects in the group are to be tracked. Adding scale invariance learning to \ours will improve performance on such videos. Extending one of our primary contributions in the smart memory selective update model, additional scalability and accuracy for very long videos can be potentially achieved by using hiearchical memory structures and temporal position embedding.
\clearpage
{
    \small
    \bibliographystyle{ieeenat_fullname}
    \bibliography{main}

\begin{thebibliography}{51}
\providecommand{\natexlab}[1]{#1}
\providecommand{\url}[1]{\texttt{#1}}
\expandafter\ifx\csname urlstyle\endcsname\relax
  \providecommand{\doi}[1]{doi: #1}\else
  \providecommand{\doi}{doi: \begingroup \urlstyle{rm}\Url}\fi

\bibitem[Bekuzarov et~al.(2023)Bekuzarov, Bermudez, Lee, and Li]{Bekuzarov2023XMemPP}
Maksym Bekuzarov, Ariana Bermudez, Joon-Young Lee, and Hao Li.
\newblock { XMem++}: Production-level video segmentation from few annotated frames.
\newblock In \emph{IEEE Int. Conf. Computer Vision}, pages 635--644, 2023.

\bibitem[Brox and Malik(2010)]{Brox2010}
Thomas Brox and Jitendra Malik.
\newblock Object segmentation by long term analysis of point trajectories.
\newblock In \emph{European Conf. Computer Vision}, pages 282--295, 2010.

\bibitem[Caron et~al.(2021)Caron, Touvron, Misra, J{\'e}gou, Mairal, Bojanowski, and Joulin]{caron2021emerging}
Mathilde Caron, Hugo Touvron, Ishan Misra, Herv{\'e} J{\'e}gou, Julien Mairal, Piotr Bojanowski, and Armand Joulin.
\newblock Emerging properties in self-supervised vision transformers.
\newblock In \emph{IEEE Int. Conf. Computer Vision}, pages 9650--9660, 2021.

\bibitem[Cheng and Schwing(2022)]{cheng2022xmem}
Ho~Kei Cheng and Alexander~G Schwing.
\newblock {XMem:} long-term video object segmentation with an {Atkinson-Shiffrin} memory model.
\newblock In \emph{European Conf. Computer Vision}, pages 640--658, 2022.

\bibitem[Cheng et~al.(2021{\natexlab{a}})Cheng, Tai, and Tang]{cheng2021modular}
Ho~Kei Cheng, Yu-Wing Tai, and Chi-Keung Tang.
\newblock Modular interactive video object segmentation: Interaction-to-mask, propagation and difference-aware fusion.
\newblock In \emph{IEEE Conf. Computer Vision and Pattern Recognition}, pages 5559--5568, 2021{\natexlab{a}}.

\bibitem[Cheng et~al.(2021{\natexlab{b}})Cheng, Tai, and Tang]{cheng2021rethinking}
Ho~Kei Cheng, Yu-Wing Tai, and Chi-Keung Tang.
\newblock Rethinking space-time networks with improved memory coverage for efficient video object segmentation.
\newblock \emph{Advances in Neural Information Processing Systems}, 34:\penalty0 11781--11794, 2021{\natexlab{b}}.

\bibitem[Cheng et~al.(2024)Cheng, Oh, Price, Lee, and Schwing]{cheng2024putting}
Ho~Kei Cheng, Seoung~Wug Oh, Brian Price, Joon-Young Lee, and Alexander Schwing.
\newblock Putting the object back into video object segmentation.
\newblock In \emph{IEEE Conf. Computer Vision and Pattern Recognition}, pages 3151--3161, 2024.

\bibitem[Dosovitskiy et~al.(2020)Dosovitskiy, Beyer, Kolesnikov, Weissenborn, Zhai, Unterthiner, Dehghani, Minderer, Heigold, Gelly, et~al.]{dosovitskiy2020image}
Alexey Dosovitskiy, Lucas Beyer, Alexander Kolesnikov, Dirk Weissenborn, Xiaohua Zhai, Thomas Unterthiner, Mostafa Dehghani, Matthias Minderer, Georg Heigold, Sylvain Gelly, et~al.
\newblock An image is worth 16x16 words: Transformers for image recognition at scale.
\newblock \emph{arXiv preprint arXiv:2010.11929}, 2020.

\bibitem[Dosovitskiy et~al.(2021)Dosovitskiy, Beyer, Kolesnikov, Weissenborn, Zhai, Unterthiner, Dehghani, Minderer, Heigold, Gelly, Uszkoreit, and Houlsby]{dosovitskiy2021an}
Alexey Dosovitskiy, Lucas Beyer, Alexander Kolesnikov, Dirk Weissenborn, Xiaohua Zhai, Thomas Unterthiner, Mostafa Dehghani, Matthias Minderer, Georg Heigold, Sylvain Gelly, Jakob Uszkoreit, and Neil Houlsby.
\newblock An image is worth 16x16 words: Transformers for image recognition at scale.
\newblock In \emph{International Conference on Learning Representations}, 2021.

\bibitem[He et~al.(2017)He, Gkioxari, Doll{\'a}r, and Girshick]{he2017mask}
Kaiming He, Georgia Gkioxari, Piotr Doll{\'a}r, and Ross Girshick.
\newblock Mask r-cnn.
\newblock In \emph{Proceedings of the IEEE international conference on computer vision}, pages 2961--2969, 2017.

\bibitem[Hong et~al.(2023)Hong, Chen, Liu, Zhang, Guo, Chen, and Zhang]{hong2023lvos}
Lingyi Hong, Wenchao Chen, Zhongying Liu, Wei Zhang, Pinxue Guo, Zhaoyu Chen, and Wenqiang Zhang.
\newblock Lvos: A benchmark for long-term video object segmentation.
\newblock In \emph{Proceedings of the IEEE/CVF International Conference on Computer Vision}, pages 13480--13492, 2023.

\bibitem[Horn and Schunck(1981)]{horn1981determining}
Berthold~KP Horn and Brian~G Schunck.
\newblock Determining optical flow.
\newblock \emph{Artificial intelligence}, 17\penalty0 (1-3):\penalty0 185--203, 1981.

\bibitem[Ke et~al.(2024)Ke, Ye, Danelljan, Tai, Tang, Yu, et~al.]{ke2024segment}
Lei Ke, Mingqiao Ye, Martin Danelljan, Yu-Wing Tai, Chi-Keung Tang, Fisher Yu, et~al.
\newblock Segment anything in high quality.
\newblock \emph{Advances in Neural Information Processing Systems (NeurIPS)}, 36, 2024.

\bibitem[Kirillov et~al.(2023)Kirillov, Mintun, Ravi, Mao, Rolland, Gustafson, Xiao, Whitehead, Berg, Lo, et~al.]{kirillov2023segment}
Alexander Kirillov, Eric Mintun, Nikhila Ravi, Hanzi Mao, Chloe Rolland, Laura Gustafson, Tete Xiao, Spencer Whitehead, Alexander~C Berg, Wan-Yen Lo, et~al.
\newblock Segment anything.
\newblock In \emph{IEEE/CVF International Conference on Computer Vision (ICCV)}, pages 4015--4026, 2023.

\bibitem[Kristan et~al.(2023)Kristan, Matas, Danelljan, Felsberg, Chang, Zajc, Luke{\v{z}}i{\v{c}}, Drbohlav, Zhang, Tran, et~al.]{kristan2023first}
Matej Kristan, Ji{\v{r}}{\'\i} Matas, Martin Danelljan, Michael Felsberg, Hyung~Jin Chang, Luka~{\v{C}}ehovin Zajc, Alan Luke{\v{z}}i{\v{c}}, Ondrej Drbohlav, Zhongqun Zhang, Khanh-Tung Tran, et~al.
\newblock The first visual object tracking segmentation vots2023 challenge results.
\newblock In \emph{IEEE/CVF International Conference on Computer Vision (ICCV)}, pages 1796--1818, 2023.

\bibitem[Kristan et~al.(2024)Kristan, Matas, Tokmakov, Felsberg, \v{C}ehovin Zajc, Luke\v{z}i\v{c}, Tran, Vu, Bjorklund, Chang, and Fernandez]{Kristan2024a}
Matej Kristan, Jiri Matas, Pavel Tokmakov, Michael Felsberg, Luka \v{C}ehovin Zajc, Alan Luke\v{z}i\v{c}, Khanh-Tung Tran, Xuan-Son Vu, Johanna Bjorklund, Hyung~Jin Chang, and Gustavo Fernandez.
\newblock The second visual object tracking segmentation vots2024 challenge results, 2024.

\bibitem[Li et~al.(2013)Li, Kim, Humayun, Tsai, and Rehg]{Fuxin2013}
Fuxin Li, Taeyoung Kim, Ahmad Humayun, David Tsai, and James~M. Rehg.
\newblock Video segmentation by tracking many figure-ground segments.
\newblock In \emph{2013 IEEE International Conference on Computer Vision}, pages 2192--2199, 2013.

\bibitem[Liang et~al.(2020)Liang, Li, Jafari, and Chen]{liang2020video}
Yongqing Liang, Xin Li, Navid Jafari, and Jim Chen.
\newblock Video object segmentation with adaptive feature bank and uncertain-region refinement.
\newblock \emph{Advances in Neural Information Processing Systems}, 33:\penalty0 3430--3441, 2020.

\bibitem[Maalouf et~al.(2024)Maalouf, Jadhav, Jatavallabhula, Chahine, Vogt, Wood, Torralba, and Rus]{maalouf2024follow}
Alaa Maalouf, Ninad Jadhav, Krishna~Murthy Jatavallabhula, Makram Chahine, Daniel~M Vogt, Robert~J Wood, Antonio Torralba, and Daniela Rus.
\newblock Follow anything: Open-set detection, tracking, and following in real-time.
\newblock \emph{IEEE Robotics and Automation Letters}, 9\penalty0 (4):\penalty0 3283--3290, 2024.

\bibitem[Mahadevan and Vasconcelos(2008)]{mahadevan2008background}
Vijay Mahadevan and Nuno Vasconcelos.
\newblock Background subtraction in highly dynamic scenes.
\newblock In \emph{2008 IEEE Conference on Computer Vision and Pattern Recognition}, pages 1--6. IEEE, 2008.

\bibitem[Ma{\v{s}}ka et~al.(2023)Ma{\v{s}}ka, Ulman, Delgado-Rodriguez, G{\'o}mez-de Mariscal, Ne{\v{c}}asov{\'a}, Guerrero~Pe{\~n}a, Ren, Meyerowitz, Scherr, L{\"o}ffler, et~al.]{mavska2023cell}
Martin Ma{\v{s}}ka, Vladim{\'\i}r Ulman, Pablo Delgado-Rodriguez, Estibaliz G{\'o}mez-de Mariscal, Tereza Ne{\v{c}}asov{\'a}, Fidel~A Guerrero~Pe{\~n}a, Tsang~Ing Ren, Elliot~M Meyerowitz, Tim Scherr, Katharina L{\"o}ffler, et~al.
\newblock The cell tracking challenge: 10 years of objective benchmarking.
\newblock \emph{Nature Methods}, 20\penalty0 (7):\penalty0 1010--1020, 2023.

\bibitem[Oh et~al.(2019)Oh, Lee, Xu, and Kim]{oh2019video}
Seoung~Wug Oh, Joon-Young Lee, Ning Xu, and Seon~Joo Kim.
\newblock Video object segmentation using space-time memory networks.
\newblock In \emph{Proceedings of the IEEE/CVF international conference on computer vision}, pages 9226--9235, 2019.

\bibitem[Perazzi et~al.(2016)Perazzi, Pont-Tuset, McWilliams, {Van Gool}, Gross, and Sorkine-Hornung]{Perazzi2016}
F. Perazzi, J. Pont-Tuset, B. McWilliams, L. {Van Gool}, M. Gross, and A. Sorkine-Hornung.
\newblock A benchmark dataset and evaluation methodology for video object segmentation.
\newblock In \emph{Computer Vision and Pattern Recognition}, 2016.

\bibitem[Pont-Tuset et~al.(2017{\natexlab{a}})Pont-Tuset, Perazzi, Caelles, Arbel\'aez, Sorkine-Hornung, and {Van Gool}]{Pont-Tuset_arXiv_2017}
Jordi Pont-Tuset, Federico Perazzi, Sergi Caelles, Pablo Arbel\'aez, Alexander Sorkine-Hornung, and Luc {Van Gool}.
\newblock The 2017 davis challenge on video object segmentation.
\newblock \emph{arXiv:1704.00675}, 2017{\natexlab{a}}.

\bibitem[Pont-Tuset et~al.(2017{\natexlab{b}})Pont-Tuset, Perazzi, Caelles, Arbel{\'a}ez, Sorkine-Hornung, and Van~Gool]{pont20172017}
Jordi Pont-Tuset, Federico Perazzi, Sergi Caelles, Pablo Arbel{\'a}ez, Alex Sorkine-Hornung, and Luc Van~Gool.
\newblock The 2017 davis challenge on video object segmentation.
\newblock \emph{arXiv preprint arXiv:1704.00675}, 2017{\natexlab{b}}.

\bibitem[Radford et~al.(2021)Radford, Kim, Hallacy, Ramesh, Goh, Agarwal, Sastry, Askell, Mishkin, Clark, et~al.]{radford2021learning}
Alec Radford, Jong~Wook Kim, Chris Hallacy, Aditya Ramesh, Gabriel Goh, Sandhini Agarwal, Girish Sastry, Amanda Askell, Pamela Mishkin, Jack Clark, et~al.
\newblock Learning transferable visual models from natural language supervision.
\newblock In \emph{International conference on machine learning}, pages 8748--8763. PMLR, 2021.

\bibitem[Rahmon et~al.(2024)Rahmon, Palaniappan, Toubal, Bunyak, Rao, and Seetharaman]{rahmon2024deepftsg}
Gani Rahmon, Kannappan Palaniappan, Imad~Eddine Toubal, Filiz Bunyak, Raghuveer Rao, and Guna Seetharaman.
\newblock Deepftsg: Multi-stream asymmetric use-net trellis encoders with shared decoder feature fusion architecture for video motion segmentation.
\newblock \emph{International Journal of Computer Vision}, 132\penalty0 (3):\penalty0 776--804, 2024.

\bibitem[Ravi et~al.(2024)Ravi, Gabeur, Hu, Hu, Ryali, Ma, Khedr, R{\"a}dle, Rolland, Gustafson, et~al.]{ravi2024sam2}
Nikhila Ravi, Valentin Gabeur, Yuan-Ting Hu, Ronghang Hu, Chaitanya Ryali, Tengyu Ma, Haitham Khedr, Roman R{\"a}dle, Chloe Rolland, Laura Gustafson, et~al.
\newblock Sam 2: Segment anything in images and videos.
\newblock \emph{arXiv preprint arXiv:2408.00714}, 2024.

\bibitem[Robinson et~al.(2020)Robinson, Lawin, Danelljan, Khan, and Felsberg]{robinson2020learning}
Andreas Robinson, Felix~Jaremo Lawin, Martin Danelljan, Fahad~Shahbaz Khan, and Michael Felsberg.
\newblock Learning fast and robust target models for video object segmentation.
\newblock In \emph{Proceedings of the IEEE/CVF conference on computer vision and pattern recognition}, pages 7406--7415, 2020.

\bibitem[Ronneberger et~al.(2015)Ronneberger, Fischer, and Brox]{ronneberger2015u}
Olaf Ronneberger, Philipp Fischer, and Thomas Brox.
\newblock U-net: Convolutional networks for biomedical image segmentation.
\newblock In \emph{Medical image computing and computer-assisted intervention--MICCAI 2015: 18th international conference, Munich, Germany, October 5-9, 2015, proceedings, part III 18}, pages 234--241. Springer, 2015.

\bibitem[Shoushtarian and Bez(2005)]{shoushtarian2005practical}
Bijan Shoushtarian and Helmut~E Bez.
\newblock A practical adaptive approach for dynamic background subtraction using an invariant colour model and object tracking.
\newblock \emph{Pattern Recognition Letters}, 26\penalty0 (1):\penalty0 5--26, 2005.

\bibitem[Siam et~al.(2021)Siam, Kendall, and Jagersand]{siam2021video}
Mennatullah Siam, Alex Kendall, and Martin Jagersand.
\newblock Video class agnostic segmentation benchmark for autonomous driving.
\newblock In \emph{Proceedings of the IEEE/CVF Conference on Computer Vision and Pattern Recognition}, pages 2825--2834, 2021.

\bibitem[Tokmakov et~al.(2023)Tokmakov, Li, and Gaidon]{tokmakov2023breaking}
Pavel Tokmakov, Jie Li, and Adrien Gaidon.
\newblock Breaking the" object" in video object segmentation.
\newblock In \emph{Proceedings of the IEEE/CVF Conference on Computer Vision and Pattern Recognition}, pages 22836--22845, 2023.

\bibitem[Toubal et~al.(2023)Toubal, Al-Shakarji, Cornelison, and Palaniappan]{toubal2023ensemble}
Imad~Eddine Toubal, Noor Al-Shakarji, DDW Cornelison, and K Palaniappan.
\newblock Ensemble deep learning object detection fusion for cell tracking, mitosis, and lineage.
\newblock \emph{IEEE Open Journal of Engineering in Medicine and Biology}, 2023.

\bibitem[Tumanyan et~al.(2025)Tumanyan, Singer, Bagon, and Dekel]{tumanyan2025dino}
Narek Tumanyan, Assaf Singer, Shai Bagon, and Tali Dekel.
\newblock Dino-tracker: Taming dino for self-supervised point tracking in a single video.
\newblock In \emph{European Conference on Computer Vision}, pages 367--385. Springer, 2025.

\bibitem[Vaswani(2017)]{vaswani2017attention}
A Vaswani.
\newblock Attention is all you need.
\newblock \emph{Advances in Neural Information Processing Systems}, 2017.

\bibitem[Wang et~al.(2022)Wang, Ahsan, Li, and Hagen]{wang2022}
Yuanbo Wang, Unaiza Ahsan, Hanyan Li, and Matthew Hagen.
\newblock \emph{A Comprehensive Review of Modern Object Segmentation Approaches}.
\newblock 2022.

\bibitem[Williams and Zipser(1989)]{williams1989learning}
Ronald~J Williams and David Zipser.
\newblock A learning algorithm for continually running fully recurrent neural networks.
\newblock \emph{Neural computation}, 1\penalty0 (2):\penalty0 270--280, 1989.

\bibitem[Wu et~al.(2023)Wu, Yang, Wu, and Chan]{Wu_2023_ICCV}
Qiangqiang Wu, Tianyu Yang, Wei Wu, and Antoni~B. Chan.
\newblock Scalable video object segmentation with simplified framework.
\newblock In \emph{Proceedings of the IEEE/CVF International Conference on Computer Vision (ICCV)}, pages 13879--13889, 2023.

\bibitem[Xu et~al.(2018{\natexlab{a}})Xu, Yang, Fan, Yang, Yue, Liang, Price, Cohen, and Huang]{YoutubeVOS2018}
Ning Xu, Linjie Yang, Yuchen Fan, Jianchao Yang, Dingcheng Yue, Yuchen Liang, Brian Price, Scott Cohen, and Thomas Huang.
\newblock Youtube-vos: Sequence-to-sequence video object segmentation.
\newblock In \emph{Computer Vision -- ECCV}, pages 603--619, 2018{\natexlab{a}}.

\bibitem[Xu et~al.(2018{\natexlab{b}})Xu, Yang, Fan, Yue, Liang, Yang, and Huang]{vos2018}
Ning Xu, Linjie Yang, Yuchen Fan, Dingcheng Yue, Yuchen Liang, Jianchao Yang, and Thomas~S. Huang.
\newblock Youtube-vos: {A} large-scale video object segmentation benchmark.
\newblock \emph{CoRR}, abs/1809.03327, 2018{\natexlab{b}}.

\bibitem[Yan et~al.(2021)Yan, Peng, Fu, Wang, and Lu]{yan2021learning}
Bin Yan, Houwen Peng, Jianlong Fu, Dong Wang, and Huchuan Lu.
\newblock Learning spatio-temporal transformer for visual tracking.
\newblock In \emph{Proceedings of the IEEE/CVF international conference on computer vision}, pages 10448--10457, 2021.

\bibitem[Yang and Yang(2022)]{yang2022deaot}
Zongxin Yang and Yi Yang.
\newblock Decoupling features in hierarchical propagation for video object segmentation.
\newblock \emph{Advances in Neural Information Processing Systems (NeurIPS)}, 2022.

\bibitem[Yang et~al.(2020)Yang, Wei, and Yang]{yang2020collaborative}
Zongxin Yang, Yunchao Wei, and Yi Yang.
\newblock Collaborative video object segmentation by foreground-background integration.
\newblock In \emph{European Conference on Computer Vision}, pages 332--348. Springer, 2020.

\bibitem[Yang et~al.(2021{\natexlab{a}})Yang, Wei, and Yang]{yang2021associating}
Zongxin Yang, Yunchao Wei, and Yi Yang.
\newblock Associating objects with transformers for video object segmentation.
\newblock \emph{Advances in Neural Information Processing Systems (NeurIPS)}, 34:\penalty0 2491--2502, 2021{\natexlab{a}}.

\bibitem[Yang et~al.(2021{\natexlab{b}})Yang, Wei, and Yang]{yang2021collaborative}
Zongxin Yang, Yunchao Wei, and Yi Yang.
\newblock Collaborative video object segmentation by multi-scale foreground-background integration.
\newblock \emph{IEEE Transactions on Pattern Analysis and Machine Intelligence}, 44\penalty0 (9):\penalty0 4701--4712, 2021{\natexlab{b}}.

\bibitem[Yang et~al.(2024)Yang, Miao, Wei, Wang, Wang, and Yang]{yang2024scalable}
Zongxin Yang, Jiaxu Miao, Yunchao Wei, Wenguan Wang, Xiaohan Wang, and Yi Yang.
\newblock Scalable video object segmentation with identification mechanism.
\newblock \emph{IEEE Transactions on Pattern Analysis and Machine Intelligence}, 2024.

\bibitem[Ye et~al.(2022)Ye, Chang, Ma, Shan, and Chen]{ye2022joint}
Botao Ye, Hong Chang, Bingpeng Ma, Shiguang Shan, and Xilin Chen.
\newblock Joint feature learning and relation modeling for tracking: A one-stream framework.
\newblock In \emph{European Conference on Computer Vision}, pages 341--357. Springer, 2022.

\bibitem[Zagoruyko and Komodakis(2016)]{zagoruyko2016wide}
Sergey Zagoruyko and Nikos Komodakis.
\newblock Wide residual networks.
\newblock \emph{arXiv preprint arXiv:1605.07146}, 2016.

\bibitem[Zhou et~al.(2024)Zhou, Pang, and Wang]{Zhou2024Rmem}
Junbao Zhou, Ziqi Pang, and Yu-Xiong Wang.
\newblock { RMem: Restricted Memory Banks Improve Video Object Segmentation }.
\newblock In \emph{2024 IEEE/CVF Conference on Computer Vision and Pattern Recognition (CVPR)}, pages 18602--18611, 2024.

\bibitem[Zhu et~al.(2023)Zhu, Chen, Hao, Chang, Zhang, Wang, Lu, Luo, He, Lan, Chen, and Li]{hqtrack}
Jiawen Zhu, Zhenyu Chen, Zeqi Hao, Shijie Chang, Lu Zhang, Dong Wang, Huchuan Lu, Bin Luo, Jun-Yan He, Jin-Peng Lan, Hanyuan Chen, and Chenyang Li.
\newblock Tracking anything in high quality, 2023.

\end{thebibliography}
}

\end{document}